\documentclass[runningheads]{llncs}

\usepackage{eccv}

\usepackage{eccvabbrv}

\usepackage{graphicx}
\usepackage{booktabs}
\usepackage{multirow}
\usepackage{xcolor}
\newcommand{\red}[1]{\textcolor{black}{#1}}
\newcommand{\blue}[1]{\textcolor{black}{#1}}
\newcommand{\olive}[1]{\textcolor{black}{#1}}

\newcommand{\brown}[1]{\textcolor{black}{#1}}
\newcommand{\shadow}[1]{}

\usepackage[accsupp]{axessibility}  

\usepackage{hyperref}

\usepackage{orcidlink}

\begin{document}
\def\r{\red}
\def\b{\blue}
\def\o{\olive}
\def\s{\shadow}
\def\n{\brown}
\title{Collision-Aware Vision-Language Learning for End-to-End Driving with Multimodal Infraction Datasets} 

\titlerunning{VLAAD}

\author{
Alex Koran\thanks{These authors contributed equally.}\inst{1,2}\orcidlink{0009-0009-5484-5694} \and
Dimitrios Sinodinos\textsuperscript{*}\inst{1,2}\orcidlink{0009-0008-0835-1282} \and
Hadi Hojjati\textsuperscript{*}\inst{1,2}\orcidlink{0000-0001-7271-0587} \and
Takuya Nanri\inst{3} \and
Fangge Chen\inst{3} \and
Narges Armanfard\textsuperscript{$\dagger$}\inst{1,2}\orcidlink{0000-0002-5880-906X}
}

\authorrunning{A.~Koran, D.~Sinodinos, H.~Hojjati, et al.}

\institute{McGill University, Montréal, Canada \and
Mila - Québec AI Institute, Montréal, Canada \and
Nissan Motor Corporation, Yokohama, Japan\\
\textsuperscript{$\dagger$} Corresponding author: \email{narges.armanfard@mcgill.ca}}

\maketitle

\begin{abstract}
\s{High infraction rates remain the primary bottleneck for end-to-end (E2E) autonomous driving, as evidenced by the low driving scores on the CARLA Leaderboard. Despite collision-related infractions being the dominant failure mode in closed-loop evaluations, collision-aware representation learning has received limited attention. To address this gap, we introduce a framework to facilitate multimodal collision awareness in E2E driving. First, we develop a Video-Language-Augmented \r{Anomaly} Detector (VLAAD). To obtain stable and temporally localized signals, we incorporate a Multiple Instance Learning (MIL) formulation into the training procedure. We benchmark this framework using real-world videos and introduce a newly curated dataset for evaluation. Beyond standalone detection, we investigate whether these representations can enhance existing E2E models. We present CARLA-Collide, a large-scale multimodal dataset of realistic collision and infraction events collected from diverse CARLA scenarios. By integrating an auxiliary collision token into a pretrained TransFuser++ agent, we demonstrate a 14.12\% relative increase in driving score with minimal fine-tuning. Our results highlight the potential of collision-aware multimodal representations for enhancing modern autonomous driving systems.}
High infraction rates remain the primary bottleneck for end-to-end (E2E) autonomous driving, as evidenced by the low driving scores on the CARLA Leaderboard. Despite collision-related infractions being the dominant failure mode in closed-loop evaluations, collision-aware representation learning has received limited attention. To address this gap, we first develop a Video-Language-Augmented Anomaly Detector (VLAAD), leveraging a Multiple Instance Learning (MIL) formulation to obtain stable, temporally localized collision signals for proactive prediction. To transition these capabilities into closed-loop simulations, we must overcome the limitations of existing simulator datasets, which lack multimodality and are frequently restricted to simple intersection scenarios. Therefore, we introduce CARLA-Collide, a large-scale multimodal dataset capturing realistic collision events across highly diverse road networks. Trained on this diverse simulator data, VLAAD serves as a collision-aware plug-in module that can be seamlessly integrated into existing E2E driving models. By integrating our module into a pretrained TransFuser++ agent, we demonstrate a 14.12\% relative increase in driving score with minimal fine-tuning. Beyond closed-loop evaluation, we further assess the generalization capability of VLAAD in an open-loop setting using real-world driving data. To support this analysis, we introduce Real-Collide, a multimodal dataset of diverse dashcam videos paired with semantically rich annotations for collision detection and prediction. On this benchmark, despite containing only 0.6B parameters, VLAAD outperforms a multi-billion-parameter vision-language model, achieving a 23.3\% improvement in AUC.
\end{abstract}

\section{Introduction}
\label{sec:intro}

\begin{figure} [t!]
    \centering
    \includegraphics[width=\linewidth]{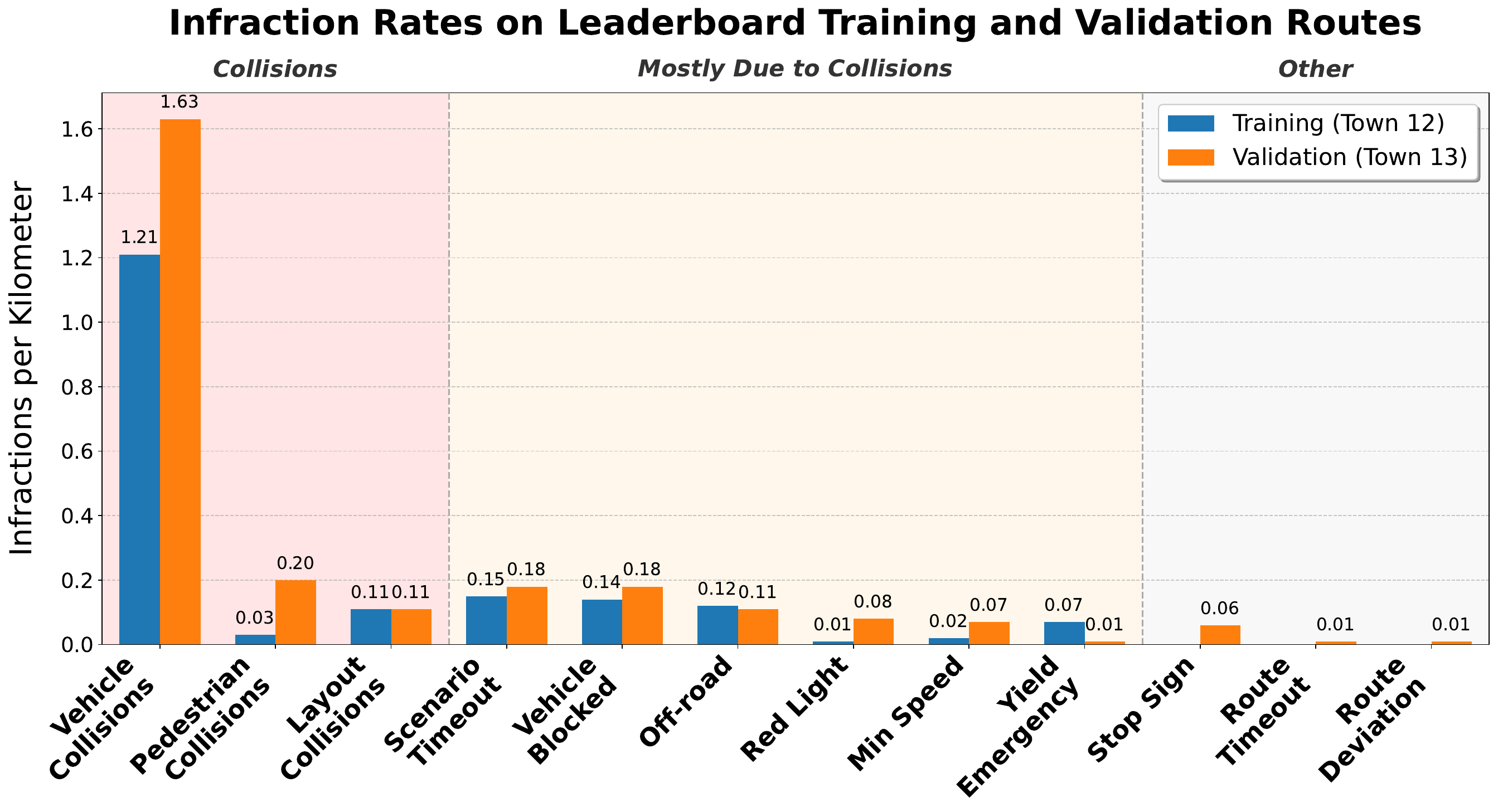}
    \caption{Infraction rates by TF++~\cite{jaeger2023hidden} on official CARLA Leaderboard training (Town 12) and validation (Town 13) routes. Collisions dominate total infractions, especially in validation. \s{EDIT the original figure by increasing the font size of the text.}}
    \label{fig:collisions}
\end{figure}

End-to-end (E2E) autonomous driving has emerged as a promising paradigm for mapping raw sensory inputs directly to driving actions, eliminating the need for modular perception–prediction–planning pipelines~\cite{chen2024end}. Using expert demonstrations, the E2E models have achieved strong offline (open-loop) performance on large-scale benchmarks such as nuScenes~\cite{caesar2020nuscenes} and Waymo Open~\cite{sun2020scalability}. However, open-loop performance has not yet been translated into reliable improvements in closed-loop driving~\cite{dauner2023parting,li2024ego}. This gap is most evident in the latest CARLA~\cite{Dosovitskiy17} Leaderboard, where top-performing models still accumulate a tremendous number of infractions and struggle to generalize to the expanded set of scenarios. As seen in Figure~\ref{fig:collisions}, a closer inspection of these \n{closed-loop} infractions reveals the central issue: collisions account for the majority of infractions in closed-loop driving evaluation. Additionally, many “non-collision” events, such as scenario timeouts and blocked agents, are direct consequences of a preceding collision. These observations highlight that collision risk remains one of the dominant bottlenecks preventing robust closed-loop performance, yet collision-aware representation learning has received surprisingly little attention in the context of E2E driving.

\n{With the growing interest in vision–language understanding for autonomous driving~\cite{Sima2024ECCV, simlingo}, multimodal modeling has emerged as a promising approach for learning representations of complex driving events. Language can describe actors, spatial relationships, and temporal interactions occurring in driving scenes, providing contextual information that complements visual signals extracted from video. Building on this observation, we introduce VLAAD (Video-Language-Augmented Anomaly Detector), a lightweight vision–language model designed for collision detection in autonomous driving. Here, anomaly refers specifically to collision-related events and safety-critical driving incidents. VLAAD leverages multimodal cues derived from driving video clips and their associated textual descriptions to detect collision events arising from the diverse and challenging scenarios present in the CARLA Leaderboard.}

\s{\blue{\r{To benchmark VLAAD's collision detection}, we curate a novel multimodal dataset, Real-Collide, consisting of diverse publicly available real-world dash cam videos, along with manually generated semantically rich annotations. Although we achieve promising post-hoc detection performance,}} 
\n{A practical collision detector should be able to localize collisions, and ideally to provide a predictive signal. We illustrate} that naïvely training a VLM for collision detection often produces temporally diffuse and unstable logits, making it difficult to obtain a reliable signal around the moment of impact. To address this, we \n{propose to} incorporate multiple instance learning (MIL)~\cite{lv2023unbiased} into VLAAD’s training procedure. MIL encourages the model to concentrate positive evidence within short, high-salience segments, yielding more precise and stable transient responses to collision onset. This formulation provides collision-aware representations that are sharper, more temporally localized, and more suitable for downstream use.

\blue{To determine if these temporally localized representations can be leveraged for \s{enhanced }closed-loop driving performance, we need to evaluate our framework in the CARLA simulator. However, to effectively test in simulation, VLAAD must be trained on a multimodal collision dataset derived from CARLA. Generally speaking,} despite the controllability and realism offered by modern simulators, the options for large-scale collision representation learning in CARLA remain limited. Recent publicly available benchmarks, such as DeepAccident~\cite{wang2024deepaccident} and DiffScene~\cite{diffscene}, focus exclusively on intersection crashes and do not reflect the wide range of failure modes observed in E2E closed-loop driving. In contrast, CARLA scenarios contain collisions arising from side impacts, rear-end events, occluded actors, and much more, that occur across all parts of the road network—not just intersections. Another limitation of existing datasets is the lack of detailed language descriptions, which restricts their utility for recent VLMs explored in autonomous driving research. This motivates the development of CARLA-Collide, a dataset that captures realistic collision and infraction events from modern E2E agents across diverse towns and driving contexts, paired with automatically generated captions that describe key actors and scene elements.

\n{The proposed VLAAD serves as a collision-aware plug-in module that can be seamlessly integrated into existing end-to-end driving models. Our experiments show that, with minimal fine-tuning, VLAAD improves relative route completion (RC), infractions score (IS), and driving score (DS) by 6.62\%, 19.23\%, and 14.12\%, respectively, compared to SOTA methods like TransFuser++~\cite{jaeger2023hidden} (TF++).
Beyond closed-loop evaluation, we further assess the generalization capability of VLAAD in an open-loop setting using real-world driving data. For this purpose, we introduce Real-Collide, a multimodal dataset of diverse dashcam videos paired with semantically rich annotations for collision detection and prediction. Curated from publicly available dashcam footage, Real-Collide captures diverse real-world collision scenarios and complements existing driving datasets for evaluating multimodal collision understanding.
Despite containing only 0.6B parameters, VLAAD outperforms the multi-billion-parameter vision–language model LLaVA-Next by 23.3\% in AUC on the Real-Collide benchmark, even when the latter is fine-tuned on real-world driving data.}

Our overall contributions are summarized as follows: \textbf{(i) VLAAD}: A lightweight vision–language model that learns collision-aware representations from multimodal driving videos and, through multiple-instance learning (MIL), produces temporally localized collision signals.
\textbf{(ii) CARLA-Collide Dataset}: A large-scale multimodal dataset of realistic collision and infraction events generated by state-of-the-art driving agents across diverse CARLA Leaderboard scenarios.
\textbf{(iii) Closed-Loop Driving Integration}: A plug-in integration of VLAAD into a pretrained driving agent that improves route completion, infraction score, and driving score with minimal fine-tuning.
\textbf{(iv) Real-Collide Benchmark}: A curated multimodal benchmark of real-world dashcam videos with semantically rich annotations for evaluating collision detection in real-world driving scenarios.

\s{

\begin{enumerate}
    \item \textbf{VLAAD Collision Detector}: A lightweight VLM trained to detect collision events from the diverse scenarios in real and simulated driving videos.
    \item \textbf{MIL-Augmented Collision Representations}: A training formulation that produces stable, temporally localized collision signals.
    \item \textbf{CARLA-Collide Dataset}: A large-scale, ego-centric dataset of realistic collision and infraction events in CARLA, paired with diverse, automatically generated captions describing actors and scene context.
    \item \textbf{Downstream Driver Integration}: A simple token-level augmentation to a pretrained TF++ agent showing that collision-aware representations yield significant improvements in driving performance with minimal fine-tuning.
        \item \blue{\textbf{Real-Collide Dataset}: A novel, multimodal dataset consisting of diverse, real-world dash cam videos used to benchmark collision detection.}
\end{enumerate}}

\section{Related Work}
\label{sec:lit_review}
\n{Recent efforts have explored generating safety-critical driving scenarios in simulation. Datasets such as DeepAccident~\cite{wang2024deepaccident} and DiffScene~\cite{diffscene} synthesize accident cases in CARLA using structured templates derived from crash reports. However, these datasets focus primarily on intersection collisions and provide limited annotations, restricting their applicability to modern vision–language frameworks that require richer contextual information. Other simulation systems such as 3CSim~\cite{cavojsky20243csim} and Anovox~\cite{bogdoll2024anovox} generate anomalous driving scenarios, but the diversity of temporal collision events remains limited. In contrast, our CARLA-Collide dataset captures a broader range of collision patterns and provides multimodal annotations suitable for vision–language modeling.}

\n{Vision–language models (VLMs) provide a unified framework for interpreting complex driving scenes by aligning visual observations with semantic descriptions. Foundational models such as CLIP~\cite{CLIP} and XCLIP~\cite{XCLIP} demonstrate strong multimodal reasoning capabilities that have motivated their adoption in autonomous driving research. Recent works such as Think-Driver~\cite{ThinkDriverVLM} and SimLingo~\cite{simlingo} explore multimodal reasoning for driving decisions and scene understanding. However, these approaches either operate at slower decision-making frequencies or focus on general scene reasoning rather than explicit collision modeling. In contrast, our framework focuses on generating temporally localized collision-risk signals that can be integrated directly into end-to-end driving systems. For more detailed discussions of related works, please refer to the Appendix.}

\s{\section{Related Works}
\label{sec:lit_review}
\r{Please check whether the related work needs to be revised accordingly. [We may also move Related Work to the appendix to save space].}

\noindent\textit{Collision Data Generation.}
Recent efforts in synthetic collision data generation aim to create diverse, safety-critical scenarios in simulation. DeepAccident~\cite{wang2024deepaccident} and  DiffScene~\cite{diffscene} are representative examples that are designed to generate accident scenarios reproduced in CARLA using structured templates derived from NHTSA crash reports. These datasets are exclusively restricted to intersection-based collisions, significantly limiting the coverage of broader vehicle–pedestrian–object interactions and naturalistic driving behaviors. Furthermore, they provide only coarse annotations, making it less suitable for modern vision-language frameworks that demand fine-grained contextual cues in the form of natural language. Other simulation frameworks, such as 3CSim~\cite{cavojsky20243csim} and Anovox~\cite{bogdoll2024anovox}, focus on generating anomalous and corner cases in driving scenes. However, the temporal driving anomalies they produce are very limited, and lack the diversity required for testing real-world driving conditions.
Despite these advancements, existing simulation datasets frequently fail to span the full spectrum of collision patterns necessary for robust temporal risk modeling. Our CARLA-Collide dataset addresses this critical gap through large-scale, automatic generation of diverse collisions, enriched with multimodal annotations and VLM-ready textual descriptions.

\noindent\textit{VLMs for Road Risk Analysis.}
VLMs offer a unified, interpretable framework for understanding complex driving scenes by aligning visual content with rich textual semantics. Models such as CLIP~\cite{CLIP} and XCLIP~\cite{XCLIP} have demonstrated strong semantic reasoning capabilities, enabling robust scene understanding without the need for handcrafted perception pipelines. Several recent works have explored leveraging VLMs for driving tasks. For instance, Think-Driver~\cite{ThinkDriverVLM} uses a VLM to produce high-level driving intentions and risk assessments; however, its interaction with the simulator operates in a benchmark-style loop where commands are issued at slow, non–real-time frequencies and executed by external simulator logic rather than a fully closed-loop controller. Another recent work, SimLingo~\cite{simlingo}, is a vision-language-action (VLA) model that performs closed-loop driving in CARLA but relies on large, slow models and chain-of-thought-style reasoning. Its collision understanding emerges implicitly through its language–action alignment rather than explicit modeling of collision representations.
While these systems confirm the promise of multimodal reasoning for autonomous driving, most operate either in slow command-generation regimes or focus on general scene understanding rather than precise, early-stage collision localization. In contrast, our framework is specifically designed to generate timely, fine-grained collision risk signals over temporally extended clips. By coupling a frozen XCLIP backbone with lightweight adapters and a multiple-instance learning structure, our method explicitly models collision risk in E2E driving stacks which require fast, reliable collision indicators rather than high-latency instruction-level reasoning.}

\begin{figure}[t!]
    \centering
    \includegraphics[width=\linewidth]{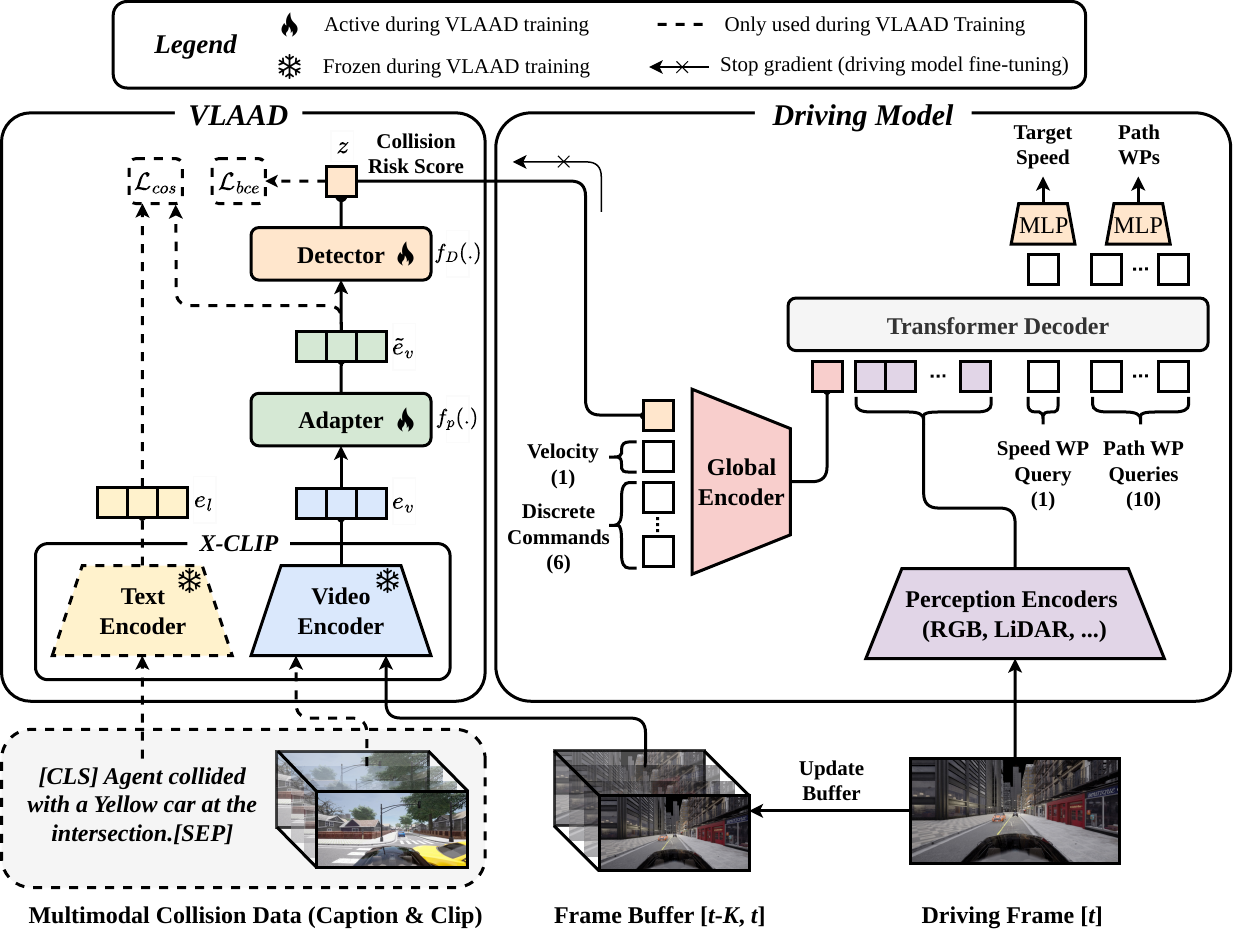}
    
    \caption{\blue{Architecture overview. \textit{VLAAD Training}: Using multimodal video-caption pairs as input, the XCLIP backbone remains frozen while the Adapter and Detector are optimized jointly using the cosine embedding loss, $\mathcal{L}_{cos}$, and binary cross-entropy loss, $\mathcal{L}_{bce}$. \textit{Driving Model Fine-Tuning \& Inference}: VLAAD operates as a frozen module, and only the video encoder is used. Its  collision risk score is concatenated with high-level states (velocity, navigation commands) and passed to the global encoder to condition the transformer decoder's waypoint and speed predictions. \red{At timestep $t$, the risk token is computed causally from the past buffer $[t-K,t]$ only (no future frames).}}}
    \label{fig:collision_detector}
\end{figure}
\section{\n{Method}}\label{sec:method}
\n{We introduce VLAAD, a lightweight vision–language model that learns collision-aware representations from multimodal driving videos and, through multiple-instance learning (MIL), produces temporally localized collision signals for downstream driving tasks.}
\subsection{VLAAD Collision Detector}
\n{To leverage the multimodal alignment capabilities of VLMs, we design a collision detector built on a frozen XCLIP backbone~\cite{XCLIP}, adapting it to the driving domain through lightweight trainable components, as illustrated in Figure~\ref{fig:collision_detector}. VLAAD retains XCLIP’s dual-encoder architecture, consisting of a video encoder and a text encoder. The video encoder processes short driving clips, while the text encoder embeds\s{automatically generated} captions describing the scene context (e.g., \textit{a car turning left at an intersection} or \textit{a vehicle collides with a pedestrian}).}

\n{To specialize the pretrained XCLIP model for collision understanding in driving scenarios, we introduce two lightweight trainable components. First, a small MLP head ($f_p$) is appended to the video encoder output to refine the visual representation ($e_v$), producing an adapted embedding ($\tilde{e}_v$) better aligned with \olive{the corresponding driving caption's embedding ($e_l$)}. Second, a collision detector ($f_D$) operates on the adapted embedding and outputs a probability $z \in [0,1]$ representing the likelihood of a collision.}

\n{The model is trained jointly using two objectives. First, a cosine embedding loss encourages semantic alignment between the adapted video embeddings and frozen text embeddings:}
\begin{equation}
\mathcal{L}_{cos}(\mathbf{\tilde{e}}_v, \mathbf{e}_l) =
\begin{cases}
1 - \cos(\mathbf{\tilde{e}}_v, \mathbf{e}_l), & \text{if } \mathbf{\tilde{e}}_v, \mathbf{e}_l \in s_i \\[6pt]
\max\left(0,\; \cos(\mathbf{\tilde{e}}_v, \mathbf{e}_l) \right), & \text{if } \mathbf{\tilde{e}}_v, \mathbf{e}_l \notin s_i
\end{cases}
\end{equation}
\n{where $\mathbf{\tilde{e}}_v, \mathbf{e}_l \in s_i$ indicates that both embeddings originate from the $i$-th sample $s_i$, and $\cos(\cdot)$ denotes cosine similarity. Second, a binary cross-entropy loss is used to distinguish collision from non-collision scenarios.}

\n{This joint training encourages the model to learn semantically aligned yet discriminative representations that capture safety-critical events. During inference, only the video encoder and collision detector are used, avoiding the computational overhead of caption processing and enabling real-time operation. By freezing the XCLIP backbone and training only lightweight adapters and classification heads, VLAAD adapts pretrained multimodal representations to collision-aware driving scenarios while maintaining strong generalization.}

\n{Although collisions occupy only a small portion of a driving clip, supervision is typically available only at the video level, making temporal localization difficult. To address this limitation, we incorporate multiple-instance learning (MIL) into the training procedure.}

\subsection{MIL-Based Temporal Localization}
\n{MIL encourages the model to concentrate prediction evidence on the few snippets that contain collision risk while keeping predictions low during normal periods. This enables the model to localize collision events within a clip and produce temporally aligned collision signals.}

\n{We represent each video as a bag and divide it into a temporally ordered sequence of short snippets (e.g., a few frames or seconds) that preserve local motion and interaction cues. For a video \(v\), this yields $T$ snippets with embeddings \(\{e_{v,1}, \ldots, e_{v,T}\}\), while only the video label \(y \in \{0,1\}\) is observed. This corresponds to the standard MIL setting in which a video is anomalous if at least one of its snippets is anomalous.}

\n{Each snippet embedding is processed using the same lightweight adapter $f_p$ (as defined in the previous section), followed by the classifier $f_D$, producing a scalar logit $z_t$ as shown in \eqref{eq:zt}.} \n{Because parameters are shared across snippets, the model learns a consistent notion of a collision-indicative snippet across videos, producing a sequence of logits \(\{z_t\}_{t=1}^T\).}
\begin{equation} \label{eq:zt}
    z_t = f_{D}\big(f_p(e_{v,t})\big), \quad t = 1, \ldots, T
\end{equation}

\n{To obtain a video-level prediction, we apply a temperature-controlled Log-Sum-Exp (LSE) pooling operator \n{as in \eqref{eq:zhbag}}, \n{where \(\gamma > 0\) controls the sharpness of the aggregation. LSE is fully differentiable, allowing multiple high-scoring snippets to contribute gradient signal while smoothly interpolating between mean-like behavior (small \(\gamma\)) and near-max behavior (large \(\gamma\)). This is beneficial when collision events extend across several adjacent snippets.}}
\begin{equation} \label{eq:zhbag}
    \hat{z}_{\text{bag}} = \frac{1}{\gamma}\!\left(\log \sum_{t=1}^{T} \exp(\gamma z_t) - \log T\right)
\end{equation}

\n{The pooled logit \(\hat{z}_{\text{bag}}\) is supervised using binary cross-entropy loss \(\mathcal{L}_{\text{bce}}\). The vision–language alignment objective is retained but adapted to the MIL setting: for positive videos, snippet-to-text similarity is reweighted according to the pooling-induced attention, encouraging the most informative snippets to align with the textual description. For negative videos, similarities are averaged across snippets to discourage uniformly high activations.}

\n{Aside from the MIL bag structure and pooling operator, all other components of VLAAD—including the backbone, optimizer, and training schedule—remain unchanged. This allows VLAAD to produce temporally localized collision signals that act as compact risk indicators for downstream driving policies.}
\begin{figure*}[t!]
    \centering
    \includegraphics[width=\linewidth]{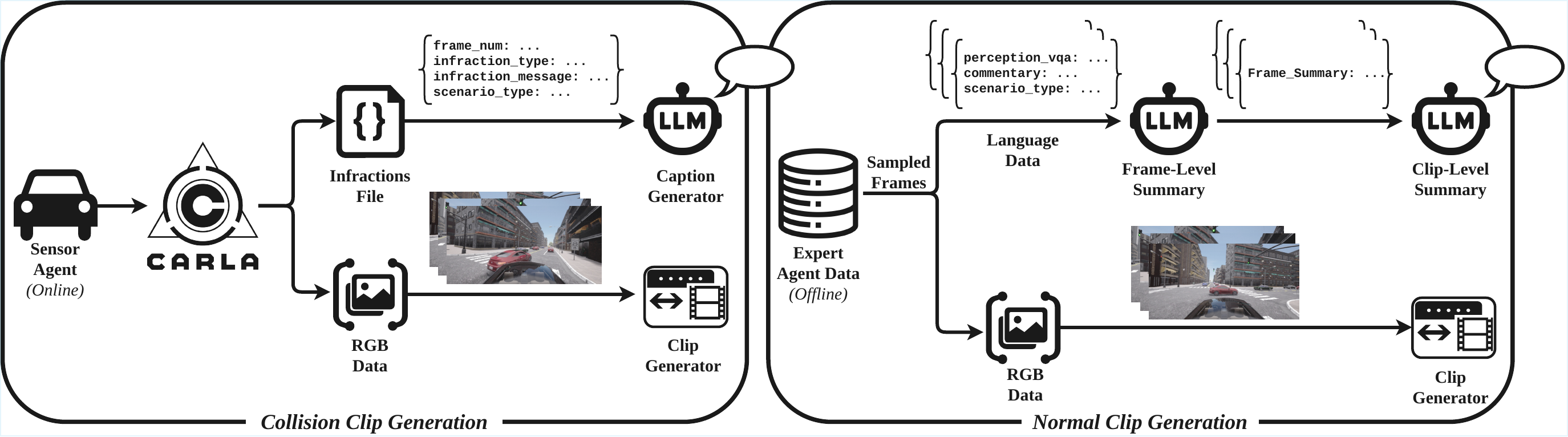}
    \caption{\b{Overview of the CARLA-Collide data generation pipeline. \textit{Collision Clips:} We record 40-frame (10s) segments from an online sensor agent (TF++), centering the collision event between 2.5s and 7.5s. Corresponding infraction logs are summarized by an LLM to produce event-specific captions. \textit{Normal Clips:} Leveraging offline expert data from SimLingo, we perform a two-stage LLM summarization—first at the frame level (incorporating VQA and scenario data) and then at the clip level—to generate cohesive driving descriptions.}}
        \label{fig:data_collection}
\end{figure*}
\n{\section{Collision Dataset Generation}
\subsection{CARLA-Collide}}

\n{Existing CARLA collision datasets~\cite{wang2024deepaccident,diffscene} are limited in scope, as they primarily focus on intersection crashes generated through hand-crafted triggers in controlled environments. In contrast, failures observed during leaderboard training and evaluation occur across diverse regions of the road network, including narrow lanes, occlusions, and mixed-traffic scenarios. To address these limitations, we introduce CARLA-Collide, a dataset that captures realistic failure cases encountered by closed-loop driving agents across the scenarios of the CARLA Leaderboard 2.0 benchmark.}
\n{Unlike prior CARLA datasets that rely on scripted collision triggers, CARLA-Collide records naturally occurring failure cases produced by state-of-the-art driving agents operating in diverse environments. As a result, the dataset reflects realistic failure modes—including occlusions, dense traffic interactions, and non-intersection collisions—that better resemble real closed-loop evaluation conditions.}
\n{To generate these failure cases, we evaluate open-source state-of-the-art driving agents such as TransFuser++~\cite{jaeger2023hidden} on the official CARLA Leaderboard. While these models achieve near-perfect performance in simple scenarios, they exhibit high infraction rates and unstable behavior in more challenging environments, making them suitable sources of realistic collision events. }

\n{We use TransFuser++ as the base driving agent and modify the official leaderboard code to enable comprehensive collision logging. Although our experiments use TransFuser++, the same data collection pipeline can be applied to other end-to-end driving agents in CARLA, allowing CARLA-Collide to capture realistic failure cases across different driving policies. Specifically, we extend the data collection functionality originally used by expert data agents to sensor-based end-to-end agents and augment it with infraction logging capabilities. Using ScenarioRunner, following~\cite{wang2024deepaccident}, we execute leaderboard training routes (Town12) and validation routes (Town13). For each run, RGB frames, control signals (throttle and steering), and infraction logs are recorded at 4 Hz, matching the sampling frequency used in TransFuser++ and SimLingo~\cite{simlingo}.}
\n{Frames are aggregated into 10-second clips (40 frames), with collisions occurring between $t=2.5$ and $t=7.5$ seconds with uniform probability. Each infraction log records: (i) frame number of the collision; (ii) infraction type (vehicle, pedestrian, or layout); (iii) infraction message describing the event; and (iv) scenario type from the CARLA Leaderboard (e.g., $\texttt{ControlLoss}$).}
\n{All infraction information is automatically generated within CARLA, enabling scalable and precise localization of collision events.}

\s{Existing collision datasets within CARLA~\cite{wang2024deepaccident,diffscene} are highly limited in scope, as they focus exclusively on intersections and use hand-crafted logic to trigger collisions in controlled scenes. In contrast, failures during leaderboard training and evaluation occur across all regions of the road network, including narrow lanes and mixed-traffic scenarios. To address these gaps, CARLA-Collide automatically captures failure cases representative of those experienced by closed-loop driving agents across the challenging scenarios of the new CARLA Leaderboard 2.0 benchmark.

Empirically, we evaluated open-source SOTA driving agents, such as TransFuser++~\cite{jaeger2023hidden}, on the CARLA Leaderboard and observed that while they achieve near-perfect performance in trivial scenarios, they exhibit high infraction rates and inconsistent performance in challenging ones. This makes them ideal candidates for generating realistic failure cases that mirror the conditions encountered during real closed-loop evaluations. Building on these observations, we leverage Transfuser++ as our base driving model and modify the official leaderboard code to enable comprehensive collision data logging. Specifically, we extend the data-collection functionality used by expert data agents to standard sensor-based end-to-end driving agents, augmenting them with infraction logging capabilities. Using ScenarioRunner, as in ~\cite{wang2024deepaccident}, we execute the leaderboard training routes (Town12) and validation routes (Town13). For each run, we record RGB frames, control measurements (throttle, steering angle), and infraction logs at 4 Hz, matching the sampling frequency of the training data used by Transfuser++ and SimLingo \cite{simlingo}. The frames are then aggregated into video clips that span 10 seconds (40 frames), with collisions occurring between $t=2.5$ and $t=7.5$ seconds with uniform probability. Each infraction log contains:
\begin{enumerate}
    \item \textbf{Frame number}: Index corresponding to the RGB frame of the collision.
    \item \textbf{Infraction type}: The type of collision (e.g., vehicle, pedestrian, or layout).
    \item \textbf{Infraction message}: Summary of the infraction (e.g., ``Agent collided against object with type=vehicle.audi.tt and id=30703 ...'').
    \item \textbf{Scenario type}: The Leaderboard scenario where the infraction occurred (e.g., $\texttt{ControlLoss}$).
\end{enumerate}
All information in the infraction log is automatically generated within CARLA, allowing for scalable and precise localization of collision events.}

\subsubsection{VLM-Enabled Captioning:}
\n{Unlike previous CARLA-based collision datasets~\cite{wang2024deepaccident,diffscene}, CARLA-Collide is designed to support vision–language modeling. Motivated by recent work showing that multimodal reasoning improves driving scene understanding~\cite{simlingo,Sima2024ECCV}, we augment both collision and normal-driving clips with textual descriptions.}
\n{For collision clips, captions are automatically derived from structured infraction logs, summarizing key actors and event dynamics. For normal-driving clips, we leverage the open-source SimLingo dataset, which contains millions of expert-driven frames annotated with perception-level VQA and commentary derived from DriveLM~\cite{Sima2024ECCV}. Frame-level annotations are aggregated into clip-level captions through a two-stage summarization process: first summarizing each frame individually, then summarizing across all frames in a clip.}
\n{Summaries are generated using lightweight open-source LLMs—Llama 3.2 (3B)~\cite{dubey2024llama} or Gemma 2 (2B)~\cite{team2024gemma}—sampled with equal probability. Each model is prompted to produce concise scene summaries from the provided annotations. This automated pipeline enables scalable generation of consistent, domain-relevant captions for both collision and normal-driving clips. Figure~\ref{fig:data_collection} illustrates the overall data collection process.}

\s{Unlike prior CARLA-based collision datasets~\cite{wang2024deepaccident,diffscene}, CARLA-Collide is designed to be VLM–ready. Motivated by recent evidence showing that multimodal reasoning enhances driving scene understanding \cite{simlingo,Sima2024ECCV}, we enrich both collision and normal-driving clips with high-quality textual descriptions. For collision clips, we automatically derive descriptive captions from the structured infraction logs, summarizing key actors, context, and event dynamics. For normal driving clips, we leverage open-sourced SimLingo data, which contains millions of expert-driven frames annotated with perception-level VQA and commentary from DriveLM \cite{Sima2024ECCV}. We aggregate these frame-level annotations into clip-level captions by performing two stages of summarization: first for each frame individually, then across all frames in a clip. For all summarization steps, we randomly query lightweight open-source LLMs, specifically Llama 3.2 (3b) ~\cite{dubey2024llama} or Gemma 2 (2b)~\cite{team2024gemma} with equal probability, and simply prompt them to provide a summary based on the provided context. This process streamlines automation and ensures consistent and domain-relevant natural language captions for all normal and collision clips. An illustration of this data collection process can be found in Figure~\ref{fig:data_collection}. Further details regarding the system prompt and LLM configurations can be found in the Appendix.}
\begin{table}[t!]
\caption{Summary of collision clip statistics in CARLA-Collide.}
\centering
\begin{tabular}{cccccccc}
\toprule
Split & \begin{tabular}[c]{@{}c@{}}Veh. \\ Coll.\end{tabular} & \begin{tabular}[c]{@{}c@{}}Ped. \\ Coll.\end{tabular} & \begin{tabular}[c]{@{}c@{}}Lay. \\ Coll.\end{tabular} & \begin{tabular}[c]{@{}c@{}}Total \\ Clips\end{tabular} & \begin{tabular}[c]{@{}c@{}}Unique \\ Scenarios\end{tabular} & Captions & Frames \\
\midrule
Training & 920 & 26 & 98 & 1044 & 37 & \checkmark & 41.76k \\
Testing & 411 & 29 & 37 & 477 & 34 & \checkmark & 19.08k \\
\midrule
Total & 1331 & 55 & 135 & 1521 & 38 & \checkmark & 60.84k \\
\bottomrule
\end{tabular}
\label{tab:collision_table}
\end{table}
\subsubsection{Dataset Composition:}
\n{CARLA-Collide contains thousands of automatically logged collision events involving vehicles, pedestrians, and static objects across diverse leaderboard scenarios. Table~\ref{tab:collision_table} summarizes dataset statistics. In addition to collision clips, we process over 2.4M frames from the SimLingo dataset to generate 46,845 training and 13,823 testing clips of normal driving behavior with corresponding captions.}
\n{Together, these clips form a large-scale benchmark of realistic collision events for training and evaluating collision-aware driving representations.}
\subsection{Real-Collide}
\n{To evaluate real-world generalization, we introduce Real-Collide, a curated multimodal dataset consisting of 1,000 dashcam video clips paired with descriptive captions. Real-Collide is constructed from publicly available dashcam videos collected from multiple online sources and curated to capture diverse real-world collision scenarios.}
\n{The dataset contains a balanced distribution of driving conditions, with 500 normal driving clips and 500 collision clips. Each clip is paired with semantically rich textual descriptions summarizing scene context and event dynamics.}
\n{Rather than prioritizing scale, Real-Collide emphasizes quality and diversity, providing a controlled benchmark for evaluating collision detection in real-world driving videos. Further details regarding data collection and annotation protocols are provided in the Appendix. Real-Collide can also be used in conjunction with existing datasets such as MM-AU~\cite{MMAU} and BDDX~\cite{bddx} to enable broader evaluation of multimodal collision understanding across diverse real-world driving scenarios.}

\section{\n{Collision-Aware Driver Integration}}
\n{To incorporate collision-aware representations into modern autonomous driving systems, we adopt an end-to-end integration strategy rather than a modular pipeline in which a standalone collision detector operates independently of the driving agent. While perception-to-control fusion has been explored in prior work, injecting an explicit collision-risk signal directly into a closed-loop driving system has received little attention. Our approach therefore provides a practical step toward risk-aware end-to-end autonomy.}

\n{As illustrated in Figure~\ref{fig:collision_detector}, this integration aligns naturally with the architecture of recent open-source closed-loop driving models such as TransFuser++ (TF++) and SimLingo. Although SimLingo extends TF++ with additional language–action alignment components, their underlying driving architectures remain largely similar. For efficiency and reproducibility, we therefore adopt TF++ as the base driving agent.}

\n{Specifically, the VLAAD collision-risk score is appended to the driver's global state representation alongside ego velocity and discrete navigation commands. This score provides a compact indicator of imminent collision risk that the driving system can leverage during decision making. By introducing this risk-aware signal, the agent can explicitly reason about predicted collision risk during action selection rather than relying solely on implicit cues in raw perceptual features. The same integration mechanism can be applied to other end-to-end driving agents that accept the collision-risk score as an additional input token.}

\n{Rather than retraining the driving model from scratch, we initialize from pretrained TF++ checkpoints and fine-tune the agent for only five epochs with the additional collision-risk feature. This lightweight integration shows that VLAAD can be incorporated into existing end-to-end driving systems with minimal architectural modification while yielding measurable improvements in closed-loop driving performance.}

\s{To demonstrate how our collision-aware representations can be seamlessly integrated into modern autonomous driving systems, we adopt an end-to-end design rather than a modular pipeline where a separate collision detector operates independently of the driver. Although perception-to-control fusion has been explored in various forms, to the best of our knowledge, prior work has not examined injecting a learned collision-prediction representation directly into a closed-loop driving policy, making this a meaningful step toward more risk-aware end-to-end autonomy. \blue{As illustrated in Figure \ref{fig:collision_detector}, this approach aligns naturally with the current landscape of state-of-the-art, open-source closed-loop models—most notably TF++ and SimLingo. While SimLingo includes extensive components for language-action alignment, stripping these away reveals an underlying driving structure almost identical to TF++. Due to this structural parity, we focus our main experiments on TF++, utilizing it as a substantially smaller and more computationally efficient base for rapid iteration. Specifically, the MIL-derived VLAAD logit acts as a global collision risk score appended alongside the driver's other global state features (ego velocity and discrete navigation commands)}. Rather than retraining from scratch, we leverage pretrained checkpoints and fine-tune for just 5 epochs with our included collision risk feature, effectively highlighting how easily VLAAD integrates into state-of-the-art end-to-end driving systems.}

\section{Experiments} \label{sec:experiments}
\s{\subsection{Evaluation Setup}}
Large-scale multi-modal data are essential for training and evaluating our proposed VLAAD framework. \s{Since XCLIP is trained on real-world data, we initially perform our experiments on existing real-world datasets to establish a proof-of-concept before shifting to the simulation domain.}The framework is implemented in Python using PyTorch and HuggingFace, with additional implementation details provided in the Appendix.

\noindent\textit{Data Sources.}
\s{no need to have these and the following as subsections-- you can just simply go to the next paragraph and have, e.g. \textbf{Data Sources.} and then write the relevant texts... same for the whole paper, such as section 4.2, and 3.2....  }
For real videos, we use MM-AU~\cite{MMAU} and BDDX~\cite{bddx} for training. Our primary test set is the proposed Real-Collide dataset, which captures diverse, naturally occurring anomalies—including collisions, lane departures, and near-miss events—under realistic driving conditions.
For simulation data, we utilize our proposed CARLA-Collide dataset. For controlled evaluation, we use the official CARLA Leaderboard validation routes in Town 13 as the held-out dataset. Using a town that was unseen during training is in line with CARLA Leaderboard's evaluation protocols. We utilize this specific, challenging set because the official Leaderboard (CARLA test server) is currently inactive, making it the highest standard for consistent evaluation. This setup comprises 20 very long routes spanning approximately 250 km in total, incorporating diverse weather conditions and scene complexity. This setup provides the necessary scalability and systematic coverage for evaluation.


\noindent\textit{Metrics.}
For \s{final}\olive{open-loop} evaluation, we focus on the collision score as the most safety-critical indicator. We report Area Under the ROC Curve (AUC) as the primary metric for its threshold-independent assessment of detection performance. We additionally include F1 score and accuracy to provide a complementary, threshold-dependent evaluation. Detailed procedures for threshold selection can be found in the Appendix. \olive{For evaluating closed-loop driving, we run all models on Town 13 routes (unseen during training) and record the kilometers (km) driven, collisions per km (Col/km), route completion (RC), infraction score (IS), and driving score (DS). We use the recently updated infraction score formula which was introduced in CARLA Leaderboard 2.1 to remove the incentive for early stopping. We perform each closed-loop evaluation 3 times and report the average of each metric. Details regarding the new infraction formula and significance testing can be found in the Appendix.}

\subsection{Closed-Loop Driving}
\blue{As detailed in Table \ref{tab:integration}, to contextualize the impact of our proposed integration, we first fine-tune the released TF++ checkpoints \n{on the training set of CARLA-Collide} without any collision-aware token to establish a fair additional baseline. This procedure yields virtually no improvement \n{on the validation routes (Town 13)}—and in fact, slightly degrades overall performance—confirming that simple fine-tuning alone does not meaningfully enhance closed-loop robustness. When we fine-tune the driver using our VLAAD token without MIL, performance degrades even further. This is consistent with the qualitative findings in Section~\ref{MIL} showing that the non-MIL logits are noisy and fail to provide a stable, localized estimate of collision risk. In stark contrast, VLAAD with MIL achieves the best performance across all recorded metrics. Most notably, it yields a 6.62\% improvement in Route Completion (RC), 19.23\% improvement in Infraction Score (IS), and a 14.12\% improvement in Driving Score (DS) relative to the original TF++ checkpoint (Baseline). Furthermore, the VLAAD-MIL agent gets blocked less often, resulting in a higher total number of kilometers (km) driven compared to the baselines. These results mirror our qualitative analysis: incorporating MIL produces sharper, more coherent transient responses to emerging collision risk, which appears to help the driving policy better contextualize expert behavior during imitation learning.}

\begin{table}[tb!]
\caption{Closed-loop evaluation on Town 13 (average of 3 runs). \s{VLAAD with MIL achieves the lowest collision rate per kilometer and significantly improves Route Completion (RC), Infraction Score (IS), and Driving Score (DS) over the baseline.}}
\centering
\begin{tabular}{clccccc}
\toprule
\multicolumn{2}{c}{Method} & km ($\uparrow$) & Col/km ($\downarrow$) & RC ($\uparrow$) & IS ($\uparrow$) & DS ($\uparrow$) \\
\midrule
\multicolumn{1}{l}{Baseline} & TF++ & 97.94 & 1.647 & 39.29 & 0.182 & 4.442 \\
\midrule
\multirow{3}{*}{Fine-tuned} & w/o VLAAD & 97.11 & 1.730 & 38.61 & 0.183 & 4.380 \\
 & w/ VLAAD & 89.86 & 1.871 & 36.18 & 0.187 & 4.440 \\
 & w/ VLAAD-MIL & \textbf{104.29} & \textbf{1.609} & \textbf{41.89} & \textbf{0.217} & \textbf{5.069} \\
\bottomrule
\end{tabular}
\label{tab:integration}
\end{table}

\subsection{Collision Detection Performance}
\s{We evaluate the proposed collision prediction model across real and simulated driving settings, focusing on its ability to leverage multimodal embeddings.}
\subsubsection{\s{Zero-Shot}Performance on Real Data:}
To evaluate different aspects of the model, we compare VLAAD with other baselines and perform several ablation studies. These results are presented in Table~\ref{tab:zero_shot}.

\noindent\textit{Comparison to Baseline.} As a baseline, we use the frozen multimodal XCLIP, which computes the cosine similarity between the video and text embeddings. Given a video clip and the text prompt, i.e., \textit{``a car accident is happening"}, the cosine similarity serves as the model’s collision score. This zero-shot approach performs near random (AUC 0.499), confirming that generic VLM features lack the temporal sensitivity required for driving anomalies. In contrast, the full VLAAD model substantially improves performance, achieving an AUC of 0.766 and consistently outperforming all ablated variants.

\noindent\textit{Comparison to Large VLMs.} We compare VLAAD with LoRA-fine-tuned LLaVA-Next~\cite{zhang2024llavanextvideo} (details in Appendix). \n{In Table \ref{tab:zero_shot}, FT denotes fine-tuning.} Despite LLaVA-Next having far more parameters, our lightweight adapter-based method achieves superior performance (0.766 vs. 0.533 AUC) and better generalization on unseen real data. This highlights that large, generic VLMs cannot be easily adapted to niche tasks, such as dangerous driving scene understanding; VLAAD bridges this gap through targeted, efficient adaptation.


\noindent\textit{Impact of Text:} To verify the importance of multimodal alignment, we remove the text modality and train a video-only variant, VLAAD (w/o Text). Performance drops substantially (from 0.766 to 0.721 AUC). This confirms that textual grounding \red{during training} provides crucial semantic structure: linguistic cues about accidents and near misses steer the model toward safety-relevant spatiotemporal features that video-only models can often overlook.

\noindent\textit{Impact of Adapter:} When we disable video embedding learning and train only a downstream anomaly detector on frozen XCLIP features (VLAAD (w/o Adapter)), performance declines (AUC 0.743). This indicates that refining the visual embedding space itself, via the adapter, is essential for capturing dynamic cues leading up to collisions.

\noindent\textit{Impact of MIL:} If we disable the MIL (VLAAD (w/o MIL)), we can see a dip in performance. This provides empirical quantitative evidence that MIL provides better supervised signal for collision detection. More details about the effects of MIL will be discussed in later sections.
Overall, the experiments demonstrate that video–text joint learning is more powerful than unimodal approaches for detecting dangerous events. The proposed collision detection module offers a lightweight yet effective pathway to adapt large foundation models for accident prediction, bridging the gap between general multimodal understanding and fine-grained driving scene risk assessment.
\begin{table}[t!]
    \caption{Performance comparison with baselines on the Real-Collide dataset. Results are averaged over 5 runs. VLAAD models are trained on MMAU + BDDX. \s{FT denotes fine-tuning.}}
    \centering
    \begin{tabular}{lcccc}
    \toprule
        Model & AUC $(\uparrow)$ & F1 $(\uparrow)$& Acc. $(\uparrow)$\\
        \midrule
         XCLIP Baseline & 0.499 & 0.640 & 0.552\\
         LlavaNext (FT) & 0.533 & 0.620 & 0.464\\
         \midrule
         VLAAD (w/o Text) & 0.721& 0.692 & 0.715 \\
         VLAAD (w/o Adapter) & 0.743 & 0.693 & 0.715 \\
         VLAAD (w/o MIL) & 0.751& 0.699 &0.721  \\
         \textbf{VLAAD (Full Model)} & \textbf{0.766} & \textbf{0.703} & \textbf{0.726} \\
         \bottomrule
    \end{tabular}
    \label{tab:zero_shot}
\end{table}

\subsubsection{Performance on Simulated Data:}
We further evaluate VLAAD on simulated environments (CARLA-Collide). Simulation provides systematically varied collision scenarios and environmental conditions that are difficult to capture in real-world datasets. The results are presented in Table~\ref{tab:simulation}.
\begin{table}[t]
\centering
\caption{\n{Performance on simulated data. Results are averaged over five runs. VLAAD models are trained on CARLA-Collide (excluding Town 13).}\s{Performance on simulation data. The reported results are the average of 5 runs. FT indicate a fine-tuned model. \olive{VLAAD models are trained with CARLA-Collide training dataset (excluding Town 13).}}}
\begin{tabular}{lcccc}
\toprule
Model & AUC $(\uparrow)$ & F1 $(\uparrow)$ & Acc. $(\uparrow)$ \\
\midrule
Baseline (w/o FT) & 0.522 & 0.549 & 0.530 \\
VLAAD (w/o MIL) & 0.655 & 0.638 & 0.611\\
VLAAD (w/ MIL)  & \textbf{ 0.672} & \textbf{0.651} & \textbf{0.642}\\
\bottomrule
\end{tabular}
\label{tab:simulation}
\end{table}
The Baseline (w/o FT) corresponds to the full VLAAD architecture (frozen XCLIP backbone, video adapter, and collision detector), where only the \n{Adapter} and \n{Detector} are trained on CARLA-Collide, and the model is evaluated on Town 13. This baseline performs poorly, achieving an AUC of 0.522, close to random. This is expected, as simulated data exhibits visual statistics—such as simplified textures and uniform lighting—that differ substantially from the real-world distributions on which XCLIP was originally trained, resulting in a significant domain mismatch.
\n{Fine-tuning the backbone is therefore essential \r{for closed-loop evaluation on simulated data}. By adapting the XCLIP visual encoder to simulated data, VLAAD learns to interpret synthetic visual cues and reduces the domain gap. This adaptation is less critical for real-world evaluation, where the data distribution more closely resembles that used during XCLIP pre-training.}

Incorporating MIL yields further improvements. By aggregating evidence across short temporal segments, MIL enables the model to capture gradual precursors to collisions that may not be evident from isolated frames. As a result, VLAAD with MIL achieves the best performance, reaching the highest AUC, F1, and balanced accuracy (AUC 0.672), confirming that temporal aggregation is crucial for robust pre-collision detection.
Overall, the simulated experiments show that (i) even with a trained \n{Adapter}, cross-domain deployment to a new synthetic domain remains unreliable; and (ii) fine-tuning the XCLIP backbone, particularly when combined with MIL, produces representations that transfer more effectively across simulated environments. This highlights the need for richer simulated infraction data, a gap addressed by our CARLA-Collide dataset.

\subsection{Qualitative Analysis of MIL}
\label{MIL}
\begin{figure}[t!]
    \centering
    \begin{subfigure}[t]{0.49\linewidth}
        \centering
        \includegraphics[width=\linewidth]{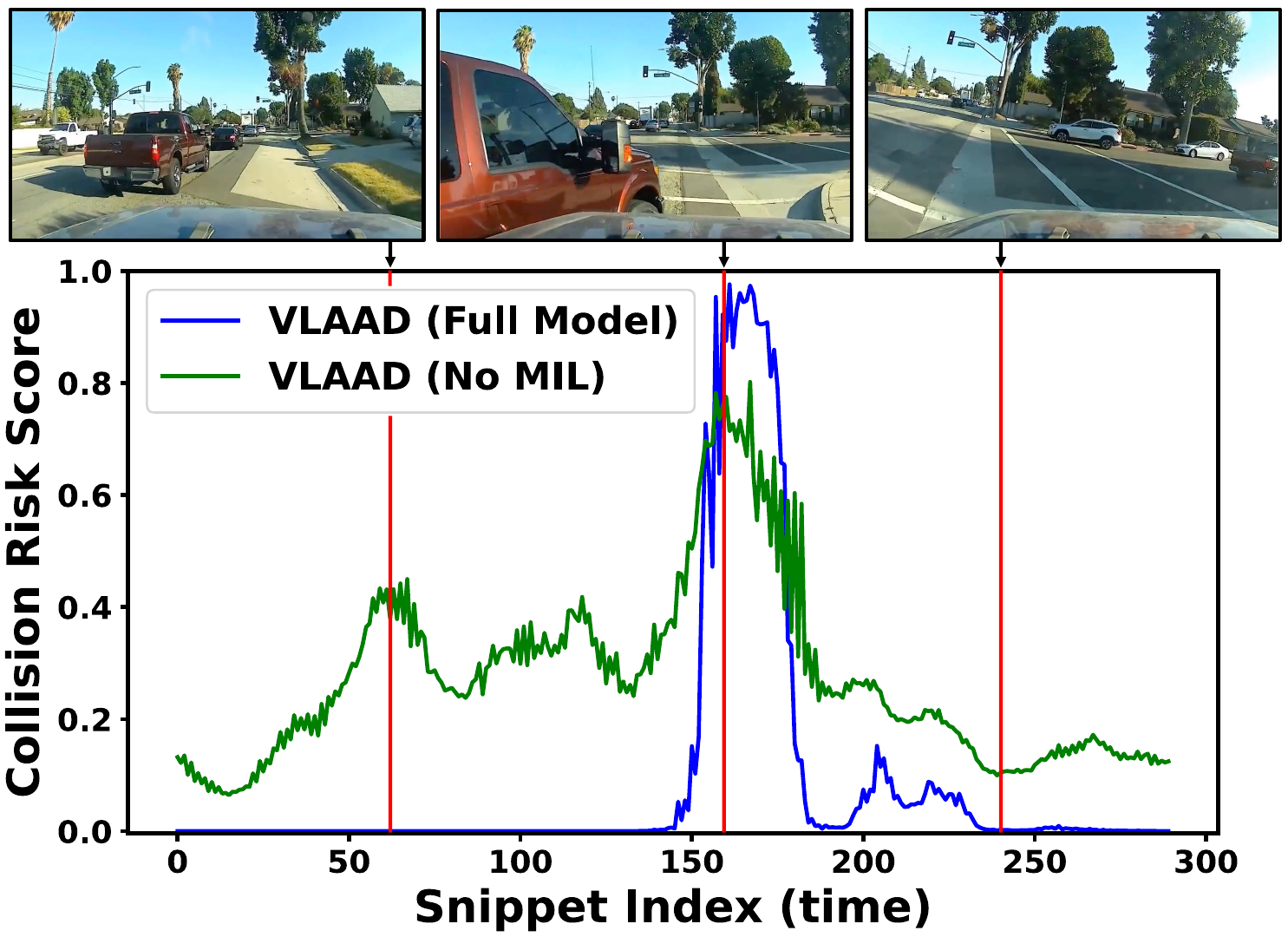}
        \caption{}
        \label{fig:mil_a}
    \end{subfigure}
    \hfill
    \begin{subfigure}[t]{0.49\linewidth}
        \centering
        \includegraphics[width=\linewidth]{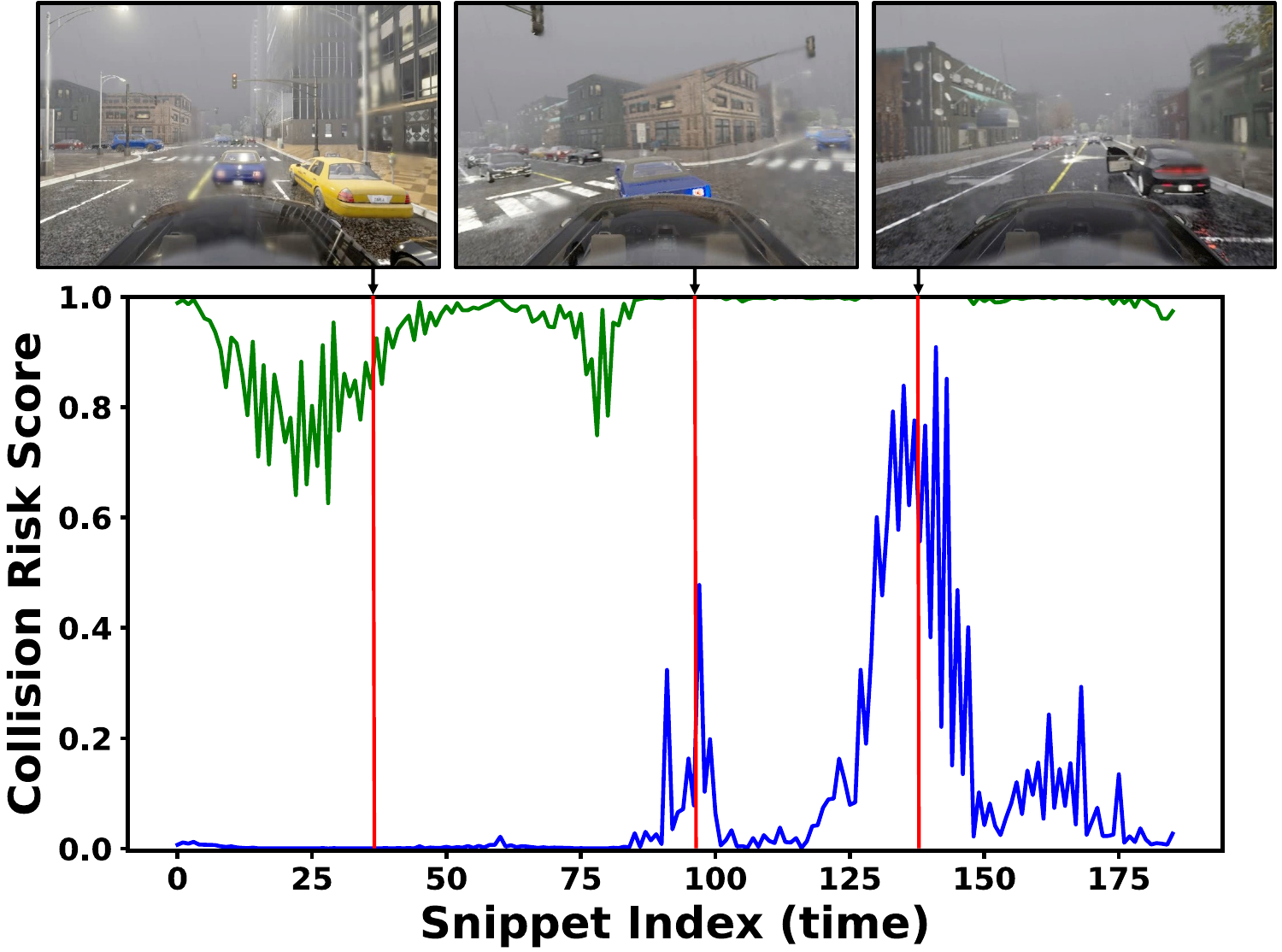}
        \caption{}
        \label{fig:mil_b}
    \end{subfigure}
   \caption{
    Temporal collision-risk predictions of VLAAD with and without MIL. \n{(a) Real-Collide clip where a pickup truck cuts in and causes a collision (evaluated on the 20\% held-out split).} \n{(b) CARLA-Collide validation clip in rain involving a minor collision and an open-door hazard (model trained on MMAU and BDDX).} 
    \s{``MIL'' produces sparse, well-localized spikes aligned with true events, while ``No MIL'' shows elevated responses across normal segments.}    }
    \label{fig:mil_combined}
\end{figure}

As shown in Tables \ref{tab:zero_shot} and \ref{tab:simulation}, clip-level AUCs improve with MIL. We examine the temporal behavior of the prediction signal to gain more insight into the snippet-level logits. These traces indicate where evidence concentrates and show that MIL yields a cleaner, time-aligned collision risk score.
In a collision example (Figure \ref{fig:mil_a}), the baseline is trained on full clips without MIL, and for visualization, both models are evaluated on the same snippets so their curves are directly comparable. MIL produces a sharp peak aligned with the collision, while the baseline yields elevated scores over a broader interval. In three normal driving examples with no collisions (Figure \ref{fig:mil_figure2}), MIL remains near-zero across each sequence, whereas the baseline shows spurious activity. Together, these results indicate that MIL provides a cleaner signal that is easier to use for early detection and for conditioning the driving policy.
\begin{figure}[t!]
  \centering
  \includegraphics[width=1\linewidth]{figures/MIL_Figure2.png}
  \caption{\n{Three normal driving clips from Real-Collide. 
VLAAD with MIL stays near zero across all videos, correctly identifying normal behavior, while the model w/o MIL shows persistent nonzero responses. 
Examples are from the held-out 20\% test split.}\s{Three normal driving clips with no collisions or accidents from the Real-Collide Dataset. ``MIL'' remains flat at zero across all videos, indicating confident recognition of normal behavior, while ``No MIL'' shows substantial nonzero activity throughout. \olive{Examples are drawn from the held-out 20\% test split of Real-Collide, and the VLAAD model is trained on the remaining 80\% of Real-Collide.}}}
  \label{fig:mil_figure2}
\end{figure}
Figure \ref{fig:mil_b} shows an example from the CARLA-Collide validation set. Here, the performance is a bit lower due to the domain shift matter discussed before. However, MIL continues to concentrate evidence near events, though absolute scores are limited by appearance and scenario differences. These \n{qualitative results show} that MIL can generate snippets that are \s{qualitatively}aligned with actual risks in the driving scenes, both in real and simulated settings.
Finally, we note that AUC is computed from one score per clip. The baseline outputs one logit per clip, and MIL pools snippet scores into one logit. If the baseline contains a high spike near the event, its pooled clip score can be close to the MIL peak, leaving the ranking of clips largely unchanged and AUCs similar. The value of MIL is the time-aligned signal that enables real-time collision prediction.



\section{Conclusion}
\label{sec:conclusion}

\s{In this work, we address the critical bottleneck of high infraction rates in closed-loop autonomous driving. Our primary contributions are the CARLA-Collide dataset, a large-scale multimodal resource for driving failures, and VLAAD, a Vision-Language-Augmented Anomaly Detection model leveraging Multiple Instance Learning for precise temporal localization. By minimally augmenting a state-of-the-art Imitation Learning agent (TF++) with our MIL-derived collision logits, we provided a proof-of-concept demonstration of the signal's practical utility, achieving a measurable reduction in closed-loop collision rates.}

\b{We introduced VLAAD, a collision-aware framework that leverages VLMs and Multiple Instance Learning (MIL) to detect and localize driving risks. To support this, we contribute Real-Collide and CARLA-Collide, two multimodal datasets dedicated to collision analysis. By integrating MIL-derived risk scores into a state-of-the-art driving policy, the pipeline achieves significant safety improvements in closed-loop evaluations with minimal overhead. Beyond these immediate results, the study provides a validated foundation for the research community to develop more robust, risk-aware representations for autonomous driving.}


%
%
\bibliographystyle{splncs04}
\bibliography{main}

@String(CVPR  = {IEEE Conf. Comput. Vis. Pattern Recog.})

@String(ECCV  = {Eur. Conf. Comput. Vis.})

@String(ICML  = {Int. Conf. Mach. Learn.})

@String(AAAI  = {AAAI})

@String(CVPR  = {CVPR})

@String(ECCV  = {ECCV})

@String(ICML  = {ICML})

@String(CVPR= {IEEE Conf. Comput. Vis. Pattern Recog.})

@String(ECCV= {Eur. Conf. Comput. Vis.})

@String(AAAI = {AAAI})

@inproceedings{Sima2024ECCV,
  title={DriveLM: Driving with Graph Visual Question Answering},
  author={Chonghao Sima and Katrin Renz and Kashyap Chitta and Li Chen and Hanxue Zhang and Chengen Xie and Jens Beißwenger and Ping Luo and Andreas Geiger and Hongyang Li},
  booktitle={Proc. of the European Conf. on Computer Vision (ECCV)},
  year={2024}
}

@inproceedings{bogdoll2024anovox,
  title={Anovox: A benchmark for multimodal anomaly detection in autonomous driving},
  author={Bogdoll, Daniel and Hamdard, Iramm and R{\"o}{\ss}ler, Lukas Namgyu and Geisler, Felix and Bayram, Muhammed and Wang, Felix and Imhof, Jan and de Campos, Miguel and Tabarov, Anushervon and Yang, Yitian and others},
  booktitle={European Conference on Computer Vision},
  pages={206--223},
  year={2024},
  organization={Springer}
}

@article{chen2024end,
  title={End-to-end autonomous driving: Challenges and frontiers},
  author={Chen, Li and Wu, Penghao and Chitta, Kashyap and Jaeger, Bernhard and Geiger, Andreas and Li, Hongyang},
  journal={IEEE Transactions on Pattern Analysis and Machine Intelligence},
  year={2024},
  publisher={IEEE}
}

@inproceedings{jaeger2023hidden,
  title={Hidden biases of end-to-end driving models},
  author={Jaeger, Bernhard and Chitta, Kashyap and Geiger, Andreas},
  booktitle={Proceedings of the IEEE/CVF International Conference on Computer Vision},
  pages={8240--8249},
  year={2023}
}

@inproceedings{caesar2020nuscenes,
  title={nuscenes: A multimodal dataset for autonomous driving},
  author={Caesar, Holger and Bankiti, Varun and Lang, Alex H and Vora, Sourabh and Liong, Venice Erin and Xu, Qiang and Krishnan, Anush and Pan, Yu and Baldan, Giancarlo and Beijbom, Oscar},
  booktitle={Proceedings of the IEEE/CVF conference on computer vision and pattern recognition},
  pages={11621--11631},
  year={2020}
}

@inproceedings{sun2020scalability,
  title={Scalability in perception for autonomous driving: Waymo open dataset},
  author={Sun, Pei and Kretzschmar, Henrik and Dotiwalla, Xerxes and Chouard, Aurelien and Patnaik, Vijaysai and Tsui, Paul and Guo, James and Zhou, Yin and Chai, Yuning and Caine, Benjamin and others},
  booktitle={Proceedings of the IEEE/CVF conference on computer vision and pattern recognition},
  pages={2446--2454},
  year={2020}
}

@inproceedings{dauner2023parting,
  title={Parting with misconceptions about learning-based vehicle motion planning},
  author={Dauner, Daniel and Hallgarten, Marcel and Geiger, Andreas and Chitta, Kashyap},
  booktitle={Conference on Robot Learning},
  pages={1268--1281},
  year={2023},
  organization={PMLR}
}

@inproceedings{li2024ego,
  title={Is ego status all you need for open-loop end-to-end autonomous driving?},
  author={Li, Zhiqi and Yu, Zhiding and Lan, Shiyi and Li, Jiahan and Kautz, Jan and Lu, Tong and Alvarez, Jose M},
  booktitle={Proceedings of the IEEE/CVF Conference on Computer Vision and Pattern Recognition},
  pages={14864--14873},
  year={2024}
}

@inproceedings{Dosovitskiy17,
  title = { {CARLA}: {An} Open Urban Driving Simulator},
  author = {Alexey Dosovitskiy and German Ros and Felipe Codevilla and Antonio Lopez and Vladlen Koltun},
  booktitle = {Proceedings of the 1st Annual Conference on Robot Learning},
  pages = {1--16},
  year = {2017}
}

@inproceedings{XCLIP,
author = {Ni, Bolin and Peng, Houwen and Chen, Minghao and Zhang, Songyang and Meng, Gaofeng and Fu, Jianlong and Xiang, Shiming and Ling, Haibin},
title = {Expanding Language-Image Pretrained Models for General Video Recognition},
year = {2022},
isbn = {978-3-031-19771-0},
publisher = {Springer-Verlag},
address = {Berlin, Heidelberg},
url = {https://doi.org/10.1007/978-3-031-19772-7_1},
doi = {10.1007/978-3-031-19772-7_1},
booktitle = {Computer Vision – ECCV 2022: 17th European Conference, Tel Aviv, Israel, October 23–27, 2022, Proceedings, Part IV},
pages = {1–18},
numpages = {18},
location = {Tel Aviv, Israel}
}

@inproceedings{CLIP,
  title={Learning Transferable Visual Models From Natural Language Supervision},
  author={Radford, Alec and Kim, Jong Wook and Hallacy, Chris and Ramesh, Aditya and Goh, Gabriel and Agarwal, Sandhini and Sastry, Girish and Askell, Amanda and Mishkin, Pamela and Clark, Jack and Krueger, Gretchen and Sutskever, Ilya},
  booktitle={Proceedings of the 38th International Conference on Machine Learning (ICML)},
  year={2021},
  organization={PMLR},
  pages={8748--8763}
}

@InProceedings{MMAU,
    author    = {Fang, Jianwu and Li, Lei-lei and Zhou, Junfei and Xiao, Junbin and Yu, Hongkai and Lv, Chen and Xue, Jianru and Chua, Tat-Seng},
    title     = {Abductive Ego-View Accident Video Understanding for Safe Driving Perception},
    booktitle = {Proceedings of the IEEE/CVF Conference on Computer Vision and Pattern Recognition (CVPR)},
    month     = {June},
    year      = {2024},
    pages     = {22030-22040}
}

@article{bddx,
  title={Textual Explanations for Self-Driving Vehicles},
  author={Kim, Jinkyu and Rohrbach, Anna and Darrell, Trevor and Canny, John and Akata, Zeynep},
  journal={Proceedings of the European Conference on Computer Vision (ECCV)},
  year={2018}
}

@InProceedings{simlingo,
  title={SimLingo: Vision-Only Closed-Loop Autonomous Driving with Language-Action Alignment},
  author={Renz, Katrin and Chen, Long and Arani, Elahe and Sinavski, Oleg},
  booktitle={Conference on Computer Vision and Pattern Recognition (CVPR)},
  year={2025}
}

@misc{zhang2024llavanextvideo,
  title={LLaVA-NeXT: A Strong Zero-shot Video Understanding Model},
  url={https://llava-vl.github.io/blog/2024-04-30-llava-next-video/},
  author={Zhang, Yuanhan and Li, Bo and Liu, haotian and Lee, Yong jae and Gui, Liangke and Fu, Di and Feng, Jiashi and Liu, Ziwei and Li, Chunyuan},
  month={April},
  year={2024}
}

@inproceedings{wang2024deepaccident,
  title={Deepaccident: A motion and accident prediction benchmark for v2x autonomous driving},
  author={Wang, Tianqi and Kim, Sukmin and Wenxuan, Ji and Xie, Enze and Ge, Chongjian and Chen, Junsong and Li, Zhenguo and Luo, Ping},
  booktitle={Proceedings of the AAAI Conference on Artificial Intelligence},
  volume={38},
  pages={5599--5606},
  year={2024}
}

@inproceedings{lv2023unbiased,
  title={Unbiased multiple instance learning for weakly supervised video anomaly detection},
  author={Lv, Hui and Yue, Zhongqi and Sun, Qianru and Luo, Bin and Cui, Zhen and Zhang, Hanwang},
  booktitle={Proceedings of the IEEE/CVF conference on computer vision and pattern recognition},
  pages={8022--8031},
  year={2023}
}

@article{dubey2024llama,
  title={The llama 3 herd of models},
  author={Dubey, Abhimanyu and Jauhri, Abhinav and Pandey, Abhinav and Kadian, Abhishek and Al-Dahle, Ahmad and Letman, Aiesha and Mathur, Akhil and Schelten, Alan and Yang, Amy and Fan, Angela and others},
  journal={arXiv e-prints},
  pages={arXiv--2407},
  year={2024}
}

@article{team2024gemma,
  title={Gemma 2: Improving open language models at a practical size},
  author={Team, Gemma and Riviere, Morgane and Pathak, Shreya and Sessa, Pier Giuseppe and Hardin, Cassidy and Bhupatiraju, Surya and Hussenot, L{\'e}onard and Mesnard, Thomas and Shahriari, Bobak and Ram{\'e}, Alexandre and others},
  journal={arXiv preprint arXiv:2408.00118},
  year={2024}
}

@INPROCEEDINGS{cavojsky20243csim,
  author={Čavojsky, Matúš and Slapak, Eugen and Dopiriak, Matúš and Bugar, Gabriel and Gazda, Juraj},
  booktitle={2024 IEEE 8th International Conference on Information and Communication Technology (CICT)}, 
  title={3CSim: CARLA Corner Case Simulation for Control Assessment in Autonomous Driving}, 
  year={2024},
  volume={},
  number={},
  pages={1-6},
  doi={10.1109/CICT64037.2024.10899666}}

@inproceedings{ThinkDriverVLM,
  author = {Zhang, Qiming and Zhu, Meixin and Yang, Hao (Frank)},
  title = {Think-Driver: From Driving-Scene Understanding to Decision-Making with Vision Language Models},
  booktitle = {Autonomous Vehicles meet Multimodal Foundation Models, ECCV 2024 Workshop},
  year = {2024}
}

@inproceedings{diffscene,
author = {Xu, Chejian and Petiushko, Aleksandr and Zhao, Ding and Li, Bo},
title = {DiffScene: diffusion-based safety-critical scenario generation for autonomous vehicles},
year = {2025},
isbn = {978-1-57735-897-8},
publisher = {AAAI Press},
url = {https://doi.org/10.1609/aaai.v39i8.32951},
doi = {10.1609/aaai.v39i8.32951},
booktitle = {Proceedings of the Thirty-Ninth AAAI Conference on Artificial Intelligence and Thirty-Seventh Conference on Innovative Applications of Artificial Intelligence and Fifteenth Symposium on Educational Advances in Artificial Intelligence},
articleno = {978},
numpages = {9},
series = {AAAI'25/IAAI'25/EAAI'25}
}

\setcounter{section}{0}
\renewcommand{\thesection}{\Alph{section}}
\renewcommand{\thesubsection}{\thesection.\arabic{subsection}}

\section{More Details on Closed-Loop Integration}
\subsection{Statement on SOTA Driving Models}
Currently, the state-of-the-art in end-to-end autonomous driving is defined by exactly two viable candidates for our experiments: TF++ and SimLingo. These are the only two fully open-sourced models that achieve true state-of-the-art performance, making them the sole practical baselines for rigorous, reproducible research. When examining SimLingo's core driving framework—also referred to by its base variant, CarLLaVA—it is evident that both models share highly comparable foundational architectures. Both rely on visual encoder(s) (with SimLingo utilizing solely images and TF++ incorporating both image and LiDAR data), a global feature encoder to process current ego velocity and navigation commands, and a transformer decoder that attends to all tokens to output path and temporal speed waypoints. The critical divergence between the two, however, lies in their computational scale and real-time viability. SimLingo is heavily burdened by a massive parameter count, a consequence not just of its supplementary language-action alignment features, but also its reliance on significantly larger encoder and decoder backbones. In contrast, TF++ utilizes a much leaner architecture without this language-alignment bloat, resulting in a model that is approximately an order of magnitude smaller. Because both models deliver very similar pure driving performance, we deliberately chose to conduct our experiments exclusively with TF++. This decision provides crucial experimental advantages: it cleanly isolates the impact of our proposed additions on the core driving model, avoiding the confounding variables and severe computational overhead of language features. Moreover, TF++ operates much closer to real-time inference than SimLingo—a priority for practical applicability—and its smaller footprint enables the rapid iteration speeds necessary for comprehensive ablation studies. Finally, it is imperative to address a known benchmarking discrepancy: previously reported results for both models on Town 14 relied on a flawed infraction score formula that inadvertently incentivized early stopping. To ensure our baseline comparisons are entirely fair and robust, our experimental setting strictly utilizes the corrected infraction score formula and evaluates the models without any early stopping techniques.

\subsection{Results on Bench2Drive}
\begin{table}[htb!]
\centering
\caption{Results on \textbf{Bench2Drive}. Scores for other methods are taken from \cite{simlingo}.}
\begin{tabular}{lcc}
\toprule
Method & Driving Score $\uparrow$ & Success Rate $\uparrow$ \\
\midrule
AD-MLP       & 18.05 & 0.00 \\
TCP          & 40.70 & 15.00 \\
VAD          & 42.35 & 15.00 \\
UniAD        & 45.81 & 16.36 \\
ThinkTwice   & 62.44 & 31.23 \\
DriveAdapter & 64.22 & 33.08 \\
SimLingo-BASE (CarLLaVA) & 85.94 & 66.82 \\
SimLingo & 85.07 & 67.27 \\
TF++ (Baseline)       & 84.21 & 67.27 \\
TF++ w/ VLAAD-MIL (Ours) & \textbf{86.97} & \textbf{71.97}\\
\midrule
\textit{PDM-Lite (Expert)} & \textit{97.02} & \textit{92.27} \\
\bottomrule
\end{tabular}
\label{tab:b2d}
\end{table}

Table \ref{tab:b2d} presents the closed-loop evaluation results on the Bench2Drive benchmark, where our proposed TF++ with VLAAD-MIL achieves state-of-the-art performance with a Driving Score of 86.97 and a Success Rate of 71.97, outperforming recent leading models like SimLingo and the standard TF++ baseline. However, it is important to contextualize these results within the specific design of Bench2Drive, which consists of 220 very short routes (approximately 150 meters) distributed across all CARLA towns, each containing exactly one safety-critical scenario. Consequently, these brief scenarios do not adequately reflect the long-term planning capabilities required to successfully navigate the extended CARLA Leaderboard routes, such as those evaluated in Town 13. Additionally, because the models are exposed to these test environments during the training phase (i.e., Town 12), Bench2Drive functions more as a "training" benchmark—reminiscent of geofenced, Level 4 autonomous driving—rather than a strict zero-shot evaluation of spatial generalization to unseen towns. Furthermore, while many earlier baselines on this benchmark were trained using the Think2Drive expert, our method, SimLingo, and the TF++ baseline leverage the significantly stronger PDM-Lite expert data (upper-bound Driving Score: 97.02). Even when building upon this robust PDM-Lite foundation, the integration of the VLAAD-MIL mechanism provides a clear, measurable boost to both the driving score and overall success rate in these dense, safety-critical scenarios.

\subsection{Training from Scratch}

\begin{table}[htb!]
\caption{Closed-loop evaluation on Town 13 (average of 3 runs). \s{VLAAD with MIL achieves the lowest collision rate per kilometer and significantly improves Route Completion (RC), Infraction Score (IS), and Driving Score (DS) over the baseline.}}
\centering
\begin{tabular}{clccccc}
\toprule
\multicolumn{2}{c}{Method} & km ($\uparrow$) & Col/km ($\downarrow$) & RC ($\uparrow$) & IS ($\uparrow$) & DS ($\uparrow$) \\
\midrule
\multicolumn{1}{l}{Baseline} & TF++ & 97.94 & 1.647 & 39.29 & 0.182 & 4.442 \\
\midrule
\multirow{3}{*}{Fine-tuned} & w/o VLAAD & 97.11 & 1.730 & 38.61 & 0.183 & 4.380 \\
 & w/ VLAAD & 89.86 & 1.871 & 36.18 & 0.187 & 4.440 \\
 & w/ VLAAD-MIL & \underline{104.29} & \underline{1.609} & \underline{41.89} & \textbf{0.217} & \textbf{5.069} \\
\midrule
\multirow{1}{*}{From-scratch}
 & w/ VLAAD-MIL & \textbf{111.50} & \textbf{1.583} & \textbf{46.41} & \underline{0.199} & \underline{4.881} \\
\bottomrule
\end{tabular}
\label{tab:integration_scratch}
\end{table}

Table~\ref{tab:integration_scratch} demonstrates that training the TF++ baseline from scratch with the VLAAD-MIL signal for 30 epochs significantly enhances navigational resilience, achieving the longest distance driven (111.50 km), the highest Route Completion (46.41), and the lowest collision rate (1.583 Col/km) compared to baseline TF++ and the 5-epoch fine-tuning approaches. However, this from-scratch method struggles more with non-collision infractions, yielding a lower Infraction Score (0.199) than the fine-tuned VLAAD-MIL model (0.217), which ultimately pulls its overall Driving Score down to 4.881 compared to the fine-tuned peak of 5.069. This discrepancy suggests that the from-scratch model may suffer from causal confusion—a common issue for driving models utilizing temporal signals—when dealing with non-collision scenarios. Yet, it is fascinating that the model avoids this pitfall for collision-based infractions, allowing it to successfully leverage temporal signals where other works have largely abandoned them. While from-scratch training serves as a strong proof that the VLAAD signal minimizes catastrophic failures and stuck states, fine-tuning remains the recommended approach, as it preserves overall driving performance and manages all infraction types effectively while saving significantly on computational cost. Moving forward, further research could explore ways to optimize the from-scratch mechanism so it learns to respect non-collision traffic rules just as effectively as it avoids collisions.

\subsection{Statistical Significance Testing for Integration Results}

To rigorously assess the statistical significance of the observed improvements in Driving Score (DS), we employ a one-sided Wilcoxon signed-rank test using a route-paired design. Each of the 20 validation routes in Town 13 represents a distinct driving scenario with varying road topology, traffic density, and weather conditions. By pairing observations by route---computing the mean DS across 3 independent evaluation seeds per configuration ($n=20$)---we effectively isolate the impact of each configuration across this diverse set of conditions. We utilize the Wilcoxon test as it is non-parametric and robust to the bounded, potentially skewed distributions of driving scores. The results reveal an impressive and statistically significant advantage for our proposed VLAAD-MIL module. Specifically, TF++ finetuned with VLAAD-MIL achieves a mean DS improvement of $+0.63$ points over the pre-trained TF++ baseline ($W=162$, $p=0.016$). Furthermore, to ensure this gain is not merely a byproduct of extra training iterations, we compared TF++ finetuned with VLAAD-MIL against a baseline fine-tuned on the exact same collision data but without the VLAAD-MIL anomaly signal. Here, TF++ finetuned with VLAAD-MIL demonstrates a highly significant mean improvement of $+0.69$ DS points ($W=170$, $p=0.007$). Conversely, fine-tuning the baseline without the VLAAD-MIL module yields no measurable improvement ($\Delta\text{DS}=-0.06$, $W=98$, $p=0.608$), definitively isolating the VLAAD-MIL anomaly-aware training signal as the true catalyst for these substantial performance gains.

\subsection{Computational Efficicency Analysis}
We evaluate computational overhead using the game-time to system-time ratio reported by the CARLA simulator in synchronous mode, where each simulation tick advances by a fixed $\Delta t = 50$~ms of game time. A ratio $r < 1$ indicates the agent's processing takes longer than real time; the effective wall-clock per tick is $\Delta t / r$. Both configurations follow standard practice and use a 3-model ensemble (i.e., taking the average output of 3 concurrent driving models trained with different seeds) to ensure robust evaluation. The TF++ baseline achieves $r = 0.25$, corresponding to 200~ms per tick (5.0~Hz effective rate). Adding VLAAD-MIL reduces the ratio to $r = 0.2$, corresponding to 250~ms per tick (4.0~Hz), introducing a fixed 50~ms of latency per tick. To contextualize this overhead against a single model: a single forward pass of the core driving model takes $\sim$67~ms. Therefore, the X-CLIP video encoder with the anomaly head (50~ms) is computationally lighter than running just one driving model, making it highly efficient. The dominant computational cost remains the driving models themselves, not VLAAD-MIL.

Furthermore, while we did not implement further optimizations because this baseline overhead was not a burden during our experiments, the following straightforward adjustments could be used to hyper-optimize the VLAAD-MIL module for strict real-time deployment constraints:
\begin{enumerate}
    \item \textbf{Clip-level output caching:} The current implementation re-runs the full X-CLIP forward pass on every simulation tick (20~Hz), even though the 8-frame buffer only changes every 5th tick (4~Hz) when a new subsampled frame arrives. Caching the video embedding between buffer updates would yield an immediate $\sim$5$\times$ reduction in X-CLIP calls with zero impact on accuracy, since the input is identical.
    \item \textbf{Preprocessor tensor caching:} The HuggingFace \texttt{XCLIPProcessor} performs CPU-bound resizing and normalization for all 8 frames on every call. Caching the preprocessed \texttt{pixel\_values} tensor and only processing the newly arrived frame would eliminate the majority of this CPU overhead.
\end{enumerate}

With these optimizations, the VLAAD-MIL overhead could likely be reduced from 50~ms to under 10~ms per tick (with a 3 model ensemble). This makes VLAAD-MIL effectively free in the context of the overall agent pipeline, demonstrating that foundational architecture improvements for collision avoidance and navigational resilience can seamlessly translate into highly efficient, deployment-ready implementations.

\section{Additional MIL Qualitative Visualization}

\begin{figure}[t]
    \centering
    \includegraphics[width=0.8\linewidth]{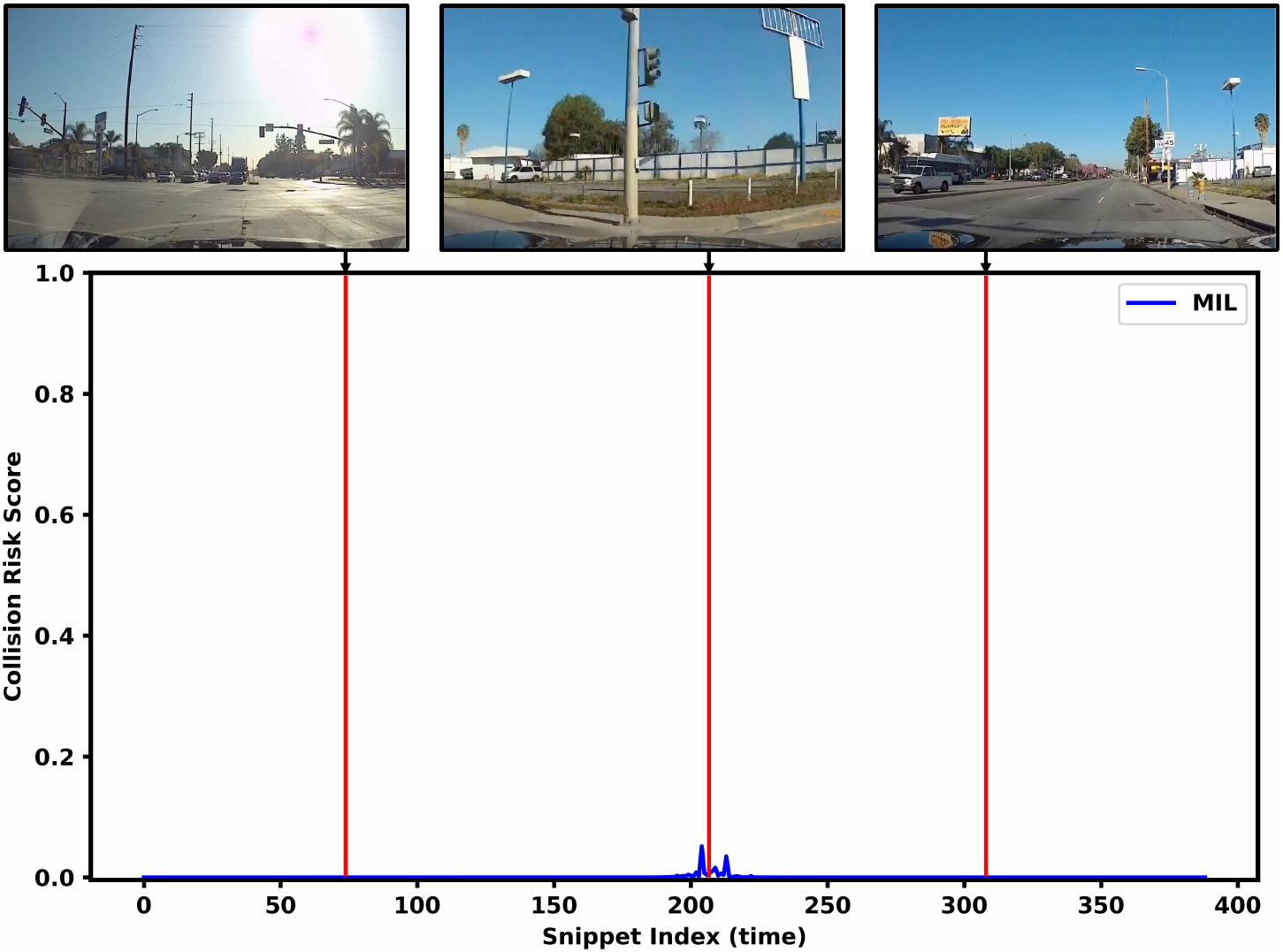}
    \caption{Temporal collision risk scores on a normal U-turn maneuver. The ego vehicle is initially stopped at a traffic light and driving nominally, then executes a left U-turn into the opposite driving direction, briefly swerving close to a utility pole in a way that visually resembles a near-crash, before completing the turn and proceeding normally. Despite the visually dramatic motion, the model maintains low snippet-level collision risk scores throughout, correctly treating the entire sequence as non-anomalous.}
    \label{fig:mil_figure_supp1}
\end{figure}

A complementary example is shown in Figure \ref{fig:mil_figure_supp1}, where MIL processes a normal clip containing a visually dramatic U-turn. The ego vehicle waits at a traffic light, then executes a sharp left U-turn that briefly brings it close to a utility pole before resuming nominal driving. MIL keeps the snippet-level collision risk score near zero for most of the clip, with only small, localized bumps around the fast turning motion. These tiny spikes are expected: they align with high steering and rapid viewpoint change, and indicate calibrated caution rather than a misclassification, since the scores remain well below the anomaly threshold and the clip is correctly treated as non-collision. This example shows that, despite being trained only with video-level labels, the MIL formulation learns a temporally selective notion of risk that respects context and scene layout, rather than simply reacting to any abrupt motion.

\begin{figure}[t]
    \centering
    \includegraphics[width=0.8\linewidth]{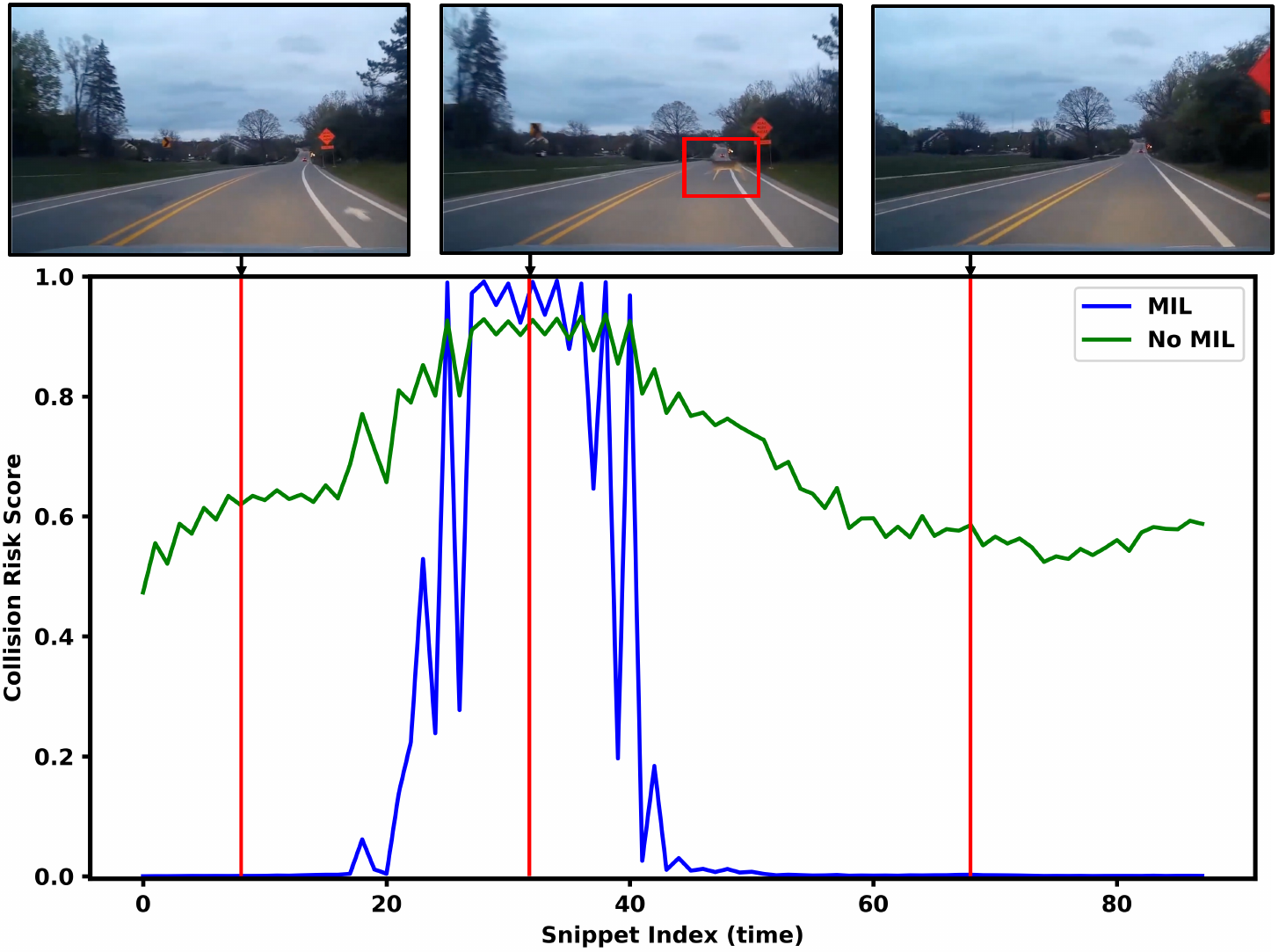}
    \caption{Temporal collision risk on a deer-crossing clip. The road is initially clear (left), a deer suddenly enters and crosses in front of the ego vehicle (middle), and the scene returns to normal once the animal exits the roadway (right). MIL keeps the collision-risk score near zero when the road is clear and produces a sharp, localized spike while the deer is on the road, whereas No MIL shows elevated scores throughout the clip, providing little temporal localization of the anomalous event.}
    \label{fig:mil_figure_supp2}
\end{figure}

Figure \ref{fig:mil_figure_supp2} shows an anomalous clip where a deer briefly crosses in front of the ego vehicle. The MIL model maintains near-zero collision-risk scores before and after the crossing and produces a narrow band of high risk only while the deer is on the road, despite being trained with video-level labels only. In contrast, the non-MIL baseline exhibits high scores over a broad portion of the clip and fails to isolate the short anomalous window. Together with Figure \ref{fig:mil_figure_supp1}, this example illustrates that MIL both suppresses false alarms on normal frames and concentrates probability mass on truly hazardous snippets, yielding a more temporally precise and interpretable risk signal.

\section{Datasets Details}
\label{sec:rationale}

\subsection{Additional Details on the Real-Collide Dataset}
\label{sec:dataset_details}

To ensure a rigorous evaluation of the VLAAD framework, we constructed a curated benchmark dataset designed to minimize domain bias and maximize scenario diversity. Unlike existing homogeneous benchmarks, Real-Collide aggregates samples from four distinct sources, each offering unique environmental characteristics and camera perspectives. The final dataset consists of $1,000$ video clips, strictly balanced between collision (positive) and normal driving (negative) scenarios.

\vspace{1mm}
\noindent\textbf{Data Sources:} The composition of the dataset is derived from the following sources:
\begin{itemize}
    \item \textbf{Nexar Collision Prediction Dataset:} We sourced high-resolution dashcam footage from the Nexar Kaggle Challenge. This dataset contributed both positive samples (collisions and near-misses) and negative samples (regular driving). Its inclusion ensures our model is tested on high-quality, consumer-grade dashcam imagery.
    \item \textbf{RetroTrucks:} To introduce domain shift and test robustness against varying ego-vehicle types, we incorporated samples from the RetroTrucks dataset. These videos are captured from a truck cabin perspective, presenting different viewing angles and motion dynamics compared to standard passenger vehicles. We utilized this source exclusively for collision/anomaly samples.
    \item \textbf{COOOL Benchmark:} The Challenge Of Out-Of-Label (COOOL) dataset provides diverse footage of road hazards and nuisance objects. We selected complex hazard scenarios from this source to populate the collision class, enriching the dataset with edge cases often missed by standard crash datasets.
    \item \textbf{YouTube Scrapes:} To cover rare accident types and fill gaps in the structured benchmarks, we scraped public dashcam footage from YouTube. These clips contributed to both the collision and normal driving partitions.
\end{itemize}

\vspace{1mm}
\noindent\textbf{Standardization and Annotation:} A critical challenge in aggregating diverse datasets is the inconsistency in label definitions (e.g., COOOL defines "hazards" broadly, while Nexar focuses on "time-to-collision"). Relying on original ground truth labels would introduce noise into the evaluation. To address this, we performed a comprehensive manual re-annotation of all $1,000$ clips. Human annotator reviewed each video to enforce a unified definition of collision risk, strictly filtering out non-threatening hazards (e.g., static debris that does not impede the ego-vehicle) to ensure a clean binary classification standard.

\begin{table}[h]
\centering
\small
\begin{tabular}{l|c|c|c}
\toprule
\textbf{Source} & \textbf{Ego-View} & \textbf{Role in Dataset} & \textbf{Scenarios} \\
\midrule
Nexar & Dashcam (Car) & Pos \& Neg & Urban, Highway \\
RetroTrucks & Dashcam (Truck) & Pos Only & Anomalies \\
COOOL & Dashcam (Car) & Pos Only & Hazards, Nuisance \\
YouTube & Various & Pos \& Neg & Diverse, Edge Cases \\
\bottomrule
\end{tabular}
\caption{Composition of the Curated Real-World Benchmark. The dataset combines standard passenger vehicle footage with truck-view anomalies and diverse web-scraped scenarios to ensure robust evaluation.}
\label{tab:curated_dataset_composition}
\end{table}

\subsection{Additional Details on the Real Datasets Used}

In this work, we conduct experiments on real data using two large-scale driving datasets designed for multimodal accident understanding and explanation generation: MM-AU (LOTVS-MMAU) and BDD-X. Both datasets provide rich, temporally aligned video–language supervision essential for studying anomaly understanding and explanation within autonomous driving systems.

MM-AU is a recently released multimodal benchmark for accident video understanding. It contains 11,727 ego-view accident videos collected from in-the-wild driving scenarios. Each video is annotated with temporally aligned textual descriptions, including accident narratives, causal explanations, and prevention strategies. MM-AU covers 58 accident categories, defined using participant-relation schemas originally introduced in DADA-2000. The dataset further provides 58,650 accident-reason answer pairs and over 2.23 million manually labeled object bounding boxes across approximately 463K frames, enabling evaluation across both perception and reasoning tasks.

Each video includes metadata such as weather, lighting, scene type, road geometry, accident windows, collision frames, and text descriptions. The annotation schema provides three aligned textual fields—Texts (accident description), Causes (why the accident occurs), and Measures (preventive advice)—which support multimodal causal reasoning. MM-AU supports a broad suite of tasks, including object detection, accident anticipation, accident reason answering, causal inference, driver attention prediction, and multimodal diffusion modeling. We use the official dataset structure provided by the authors, including the LOTVS-Cap and LOTVS-DADA partitions.

BDD-X is a video–language dataset built on the Berkeley DeepDrive driving corpus. It contains 6,970 videos totaling 77 hours of real-world driving, with each clip averaging approximately 40 seconds. The dataset provides natural language descriptions of driver actions together with explanations describing the underlying reasons for those actions. Across the dataset, annotators label over 26K action segments, each with temporal boundaries and paired (description, explanation) annotations.

The annotation protocol instructs annotators—trained and familiar with U.S. driving rules—to adopt the role of a driving instructor. For each meaningful change in behavior (e.g., slowing down, turning, merging), the annotator provides both what the driver is doing and why they are doing it, producing high-quality causal explanations tied to specific temporal segments. The dataset spans diverse environmental conditions (day/night, urban/highway/rural, various weather), providing broad coverage of everyday driving behaviors.

BDD-X serves as a clean benchmark for evaluating video-language models on causal explanation quality, grounded action understanding, and temporal video reasoning.

\subsection{CARLA-COLLIDE} 
\subsubsection{Details on Infractions Rates}
\begin{figure}[htb!]
    \centering
    \includegraphics[width=\linewidth]{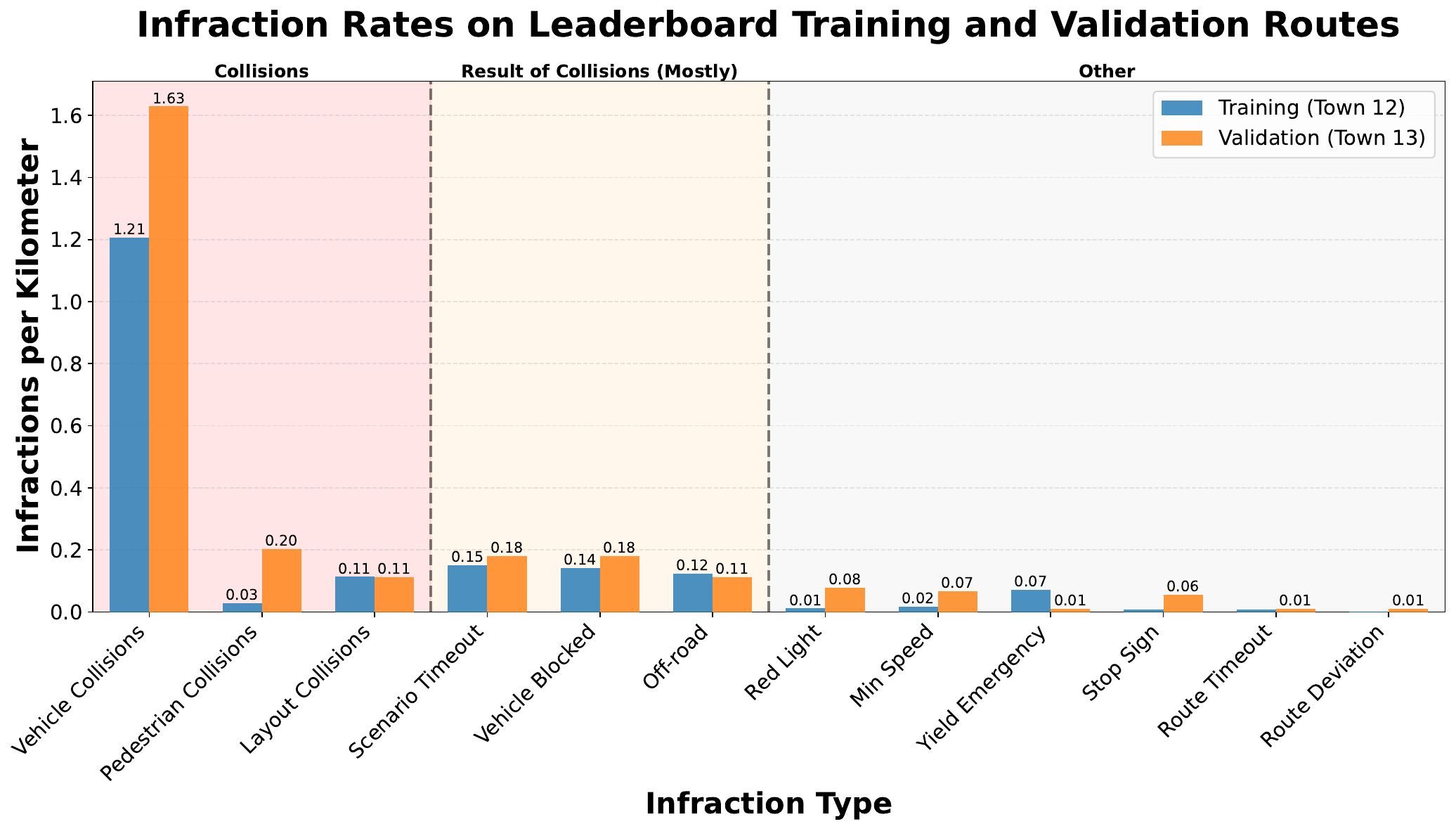}
    \caption{Infraction rates per kilometer on official CARLA leaderboard training (Town 12) and validation (Town 13) routes. Collisions dominate total infractions, especially in validation}
    \label{fig:infraction_rates_detailed}
\end{figure}

Figure \ref{fig:infraction_rates_detailed} provides a detailed look into infraction rates per kilometer between the leaderboard’s training (Town 12) and validation (Town 13) routes using an off-the-shelf TF++ driving model. Collisions are by far the most frequent infraction category, with validation routes exhibiting notably higher vehicle–vehicle collision rates (1.63 vs. 1.21) and pedestrian collisions rates (0.20 vs. 0.03) on validation routes. Non-collision infractions that are mostly consequences of collisions—such as scenario timeouts, vehicle-blocked events, and off-road violations—follow the same trend, showing slightly elevated rates in validation. These patterns indicate that when a collision occurs, agents often accumulate additional infractions as they fail to recover or complete the scenario. The remaining categories (“Other”) occur at very low frequencies. Red-light, yield-emergency, stop-sign, and route-deviation violations remain very low in comparison to collisions for both settings, suggesting that rule-based infractions are not the main drivers of overall performance degradation. Overall, the data reinforces a central motivation of our work: collisions are the primary bottleneck in closed-loop driving performance, and most harmful downstream infractions stem directly from them—highlighting the need for collision-aware learning strategies.

\subsubsection{Infractions by Scenario}

\begin{figure}[htb!]
    \centering
    \includegraphics[width=0.78\linewidth]{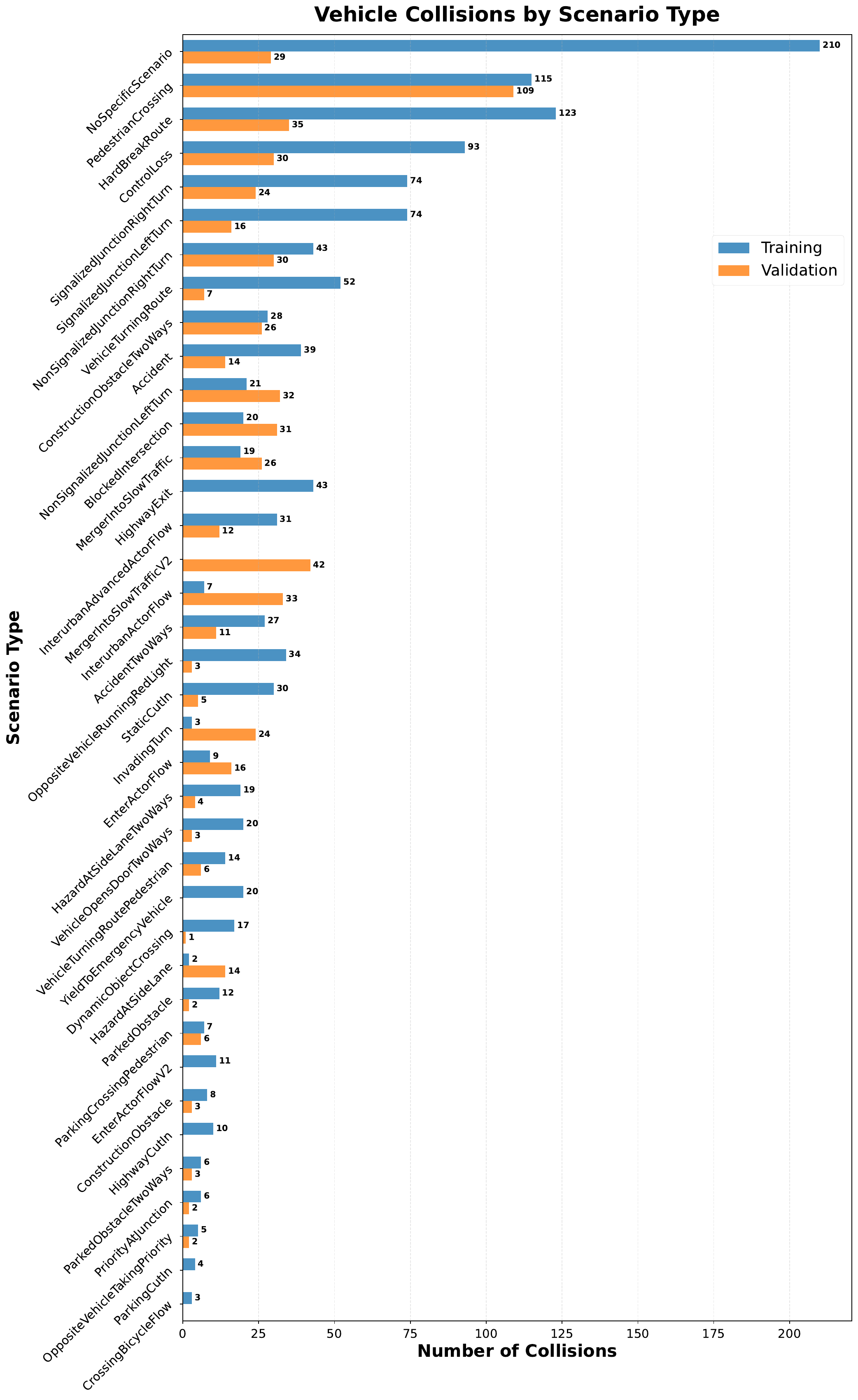}
    \caption{Vehicle collision counts across leaderboard scenario types for training (Town 12) and validation (Town 13) routes. NoSpecificScenario means the collision occurred while in transit between scenarios.}
    \label{fig:scenario_vehicle}
\end{figure}

\begin{figure}[htb!]
    \centering
    \includegraphics[width=\linewidth]{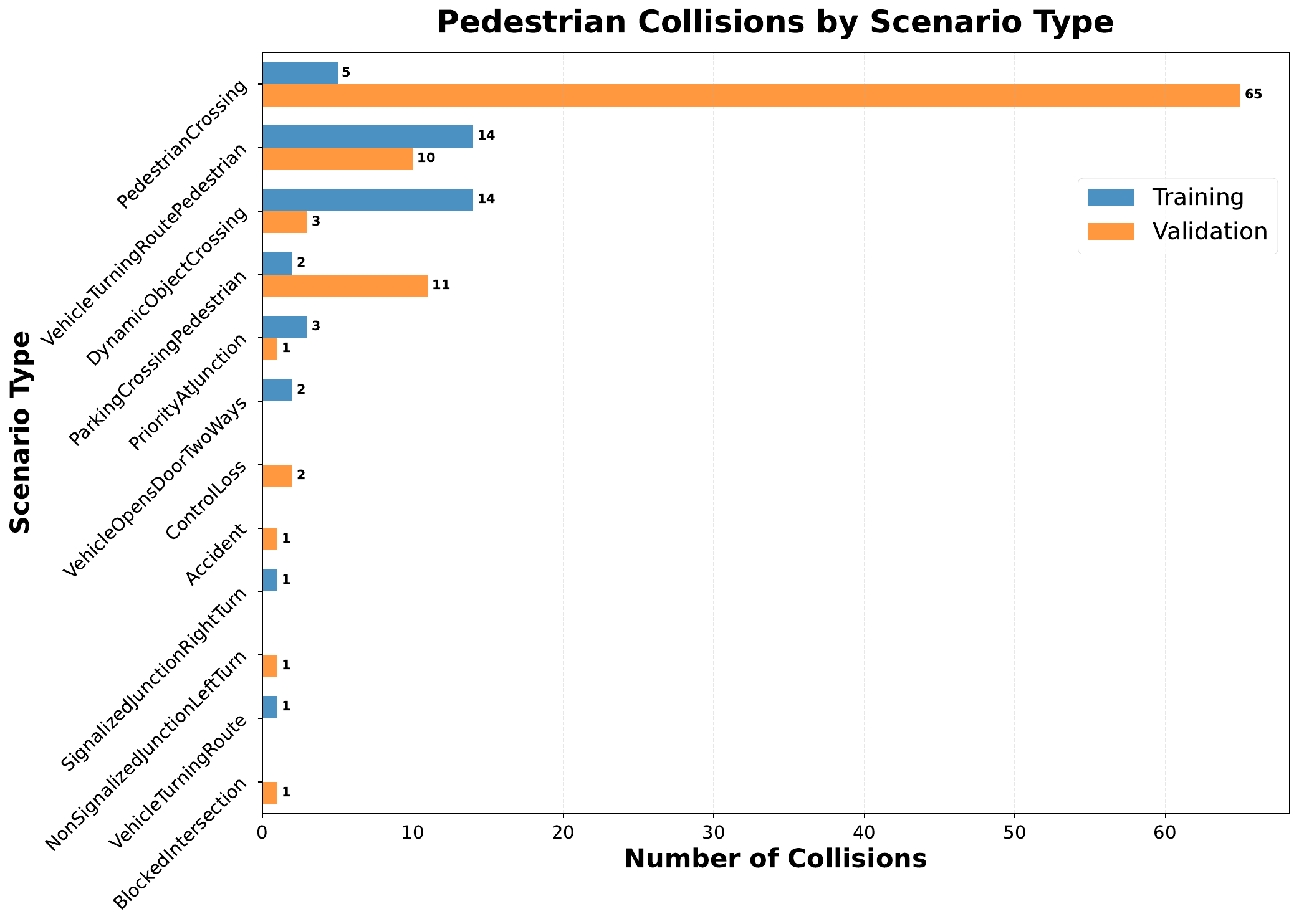}
    \caption{Pedestrian collision counts across leaderboard scenario types for training (Town 12) and validation (Town 13) routes.}
    \label{fig:scenario_pedestrian}
\end{figure}

\begin{figure}[htb!]
    \centering
    \includegraphics[width=\linewidth]{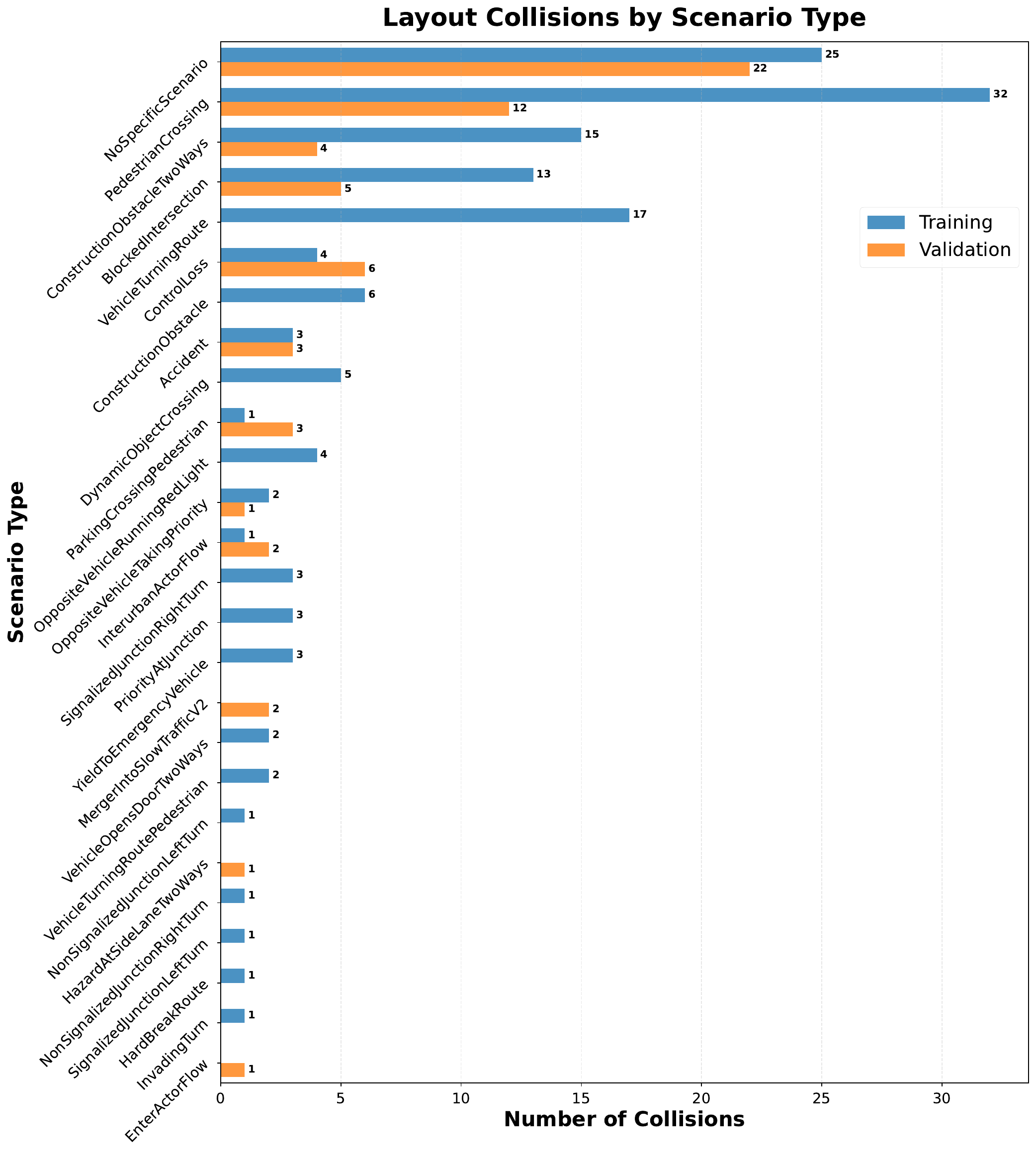}
    \caption{Layout (object) collision counts across leaderboard scenario types for training (Town 12) and validation (Town 13) routes. NoSpecificScenario means the collision occurred while in transit between scenarios.}
    \label{fig:scenario_layout}
\end{figure}

The three collision distribution plots (Figures~\ref{fig:scenario_vehicle},~\ref{fig:scenario_pedestrian},~\ref{fig:scenario_layout}) break down vehicle, pedestrian, and layout (object) collision counts by scenario type for both the training (Town 12) and validation (Town 13) leaderboard routes using a pre-trained TF++. Together, they characterize where common collisions occur in closed-loop driving, offering a detailed view of the failure landscape that informs the composition of the CARLA-Collide dataset. As a result, CARLA-Collide contains a rich and realistic mix of collision contexts that provide organic diversity essential for training collision detectors that can ultimately enable more informed, collision-aware imitation learning.

\subsection{CARLA Leaderboard Scoring System}
Driving scores on the CARLA leaderboard have historically been extremely low, with Leaderboard~2.0 agents achieving a maximum of only \textbf{6.8/100}. Under the original scoring scheme, the global driving score for the $i$-th route was computed as
\[
\text{DrivingScore}_i = R_i \, P_i,
\]
where $R_i$ denotes the percentage of route completed and $P_i$ is the infraction penalty. In Leaderboard~2.0, the infraction penalty was defined multiplicatively as
\[
P_i = \prod_{j} p_j^{\#\text{infractions}_j},
\]
where $p_j \in (0,1]$ is the penalty weight associated with infraction type $j$, and $\#\text{infractions}_j$ is the number of times infraction $j$ occurred. This multiplicative formulation, combined with the high frequency of infractions, often produced an unintuitive optimization landscape in which agents benefitted from \emph{early stopping} to avoid accumulating additional penalties~\cite{jaeger2023hidden}. Leaderboard~2.1 revises the infraction penalty to strongly incentivize \textbf{full route completion}. The updated infraction penalty is now defined as
\[
P_i = \frac{1}{1 + \sum_{j} c_j \cdot \#\text{infractions}_j},
\]
where $c_j$ denotes the severity coefficient of infraction type $j$ (e.g., $1.0$ for pedestrian collisions, $0.70$ for vehicle collisions, and $0.60$ for static-object collisions). Agents begin with an ideal score of $P_i = 1.0$, which decreases monotonically as infractions accumulate. Even under this new formulation, collisions remain doubly detrimental. First, they carry the highest penalty coefficients and therefore strongly reduce the infraction term $P_i$. Second, collisions frequently result in \emph{scenario timeouts}, \emph{route blockage}, or other cascading failures that reduce $R_i$, the route-completion term. Consequently, achieving competitive performance on Leaderboard~2.1 requires reducing collisions not only to decrease infraction penalties but also to maintain uninterrupted route progression. Therefore, improving collision robustness remains central to maximizing the overall driving score under the updated CARLA evaluation protocol.

\section{Implementation Details}

\subsection{Carla-Collide Data Generation}
\label{sec:llm_prompts}

Since the number of normal samples is significantly larger, we employ a data augmentation strategy to increase the number of collision samples. Specifically, instead of fixing the collision frame at the center of the clip, we randomly sample a position between 0.1 and 0.9 of the temporal length and place the collision frame at that location. This process is repeated five times for each sample, resulting in a fivefold increase in the number of collision samples.

As described in the main paper, we leverage lightweight open-source Large Language Models (LLMs) to generate natural language descriptions for both collision and normal driving clips. We utilize the \texttt{ollama} library to interface with two specific models: \texttt{llama3.2:3b} and \texttt{gemma2:2b}.

To ensure diversity in the generated captions, for every annotation task, we randomly sample one of these two models with uniform probability ($p=0.5$). We employ two distinct prompt templates depending on the stage of the pipeline:

\begin{itemize}
    \item \textbf{Summarization Prompt:} Used for aggregating frame-level perception data or infraction logs into concise summaries.
    \begin{quote}
        \small
        ``Summarize the following text:\\
        \textit{\{input\_text\}}\\
        Only output the summarized message with nothing before it.''
    \end{quote}
    
    \item \textbf{Paraphrasing Prompt:} Used to refine captions while maintaining domain relevance. We enforce a lexical constraint to ensure the context of driving is preserved.
    \begin{quote}
        \small
        ``Paraphrase the following text while keeping the original meaning:\\
        \textit{\{input\_text\}}\\
        Only output the paraphrased message with nothing before it. The word drive must be in your response.''
    \end{quote}
\end{itemize}

The outputs from these prompts are directly used as the textual ground truth.

\subsection{VLAAD Implementation Details}
\label{sec:implementation_details}

We implement our framework using PyTorch. All experiments are reproducible with a fixed random seed. Below we detail the specific architectural choices and training hyperparameters used for VLAAD and its Multiple Instance Learning (MIL) variant.

\vspace{1mm}
\noindent\textbf{Architecture.}
Our model utilizes a frozen XCLIP backbone. The Video Adapter projects the input visual embeddings (dimension $D=768$) to a hidden feature space of dimension $H=256$. For the MIL formulation, the model processes inputs as bags of snippets, retaining the same embedding dimensions.

\vspace{1mm}
\noindent\textbf{Training Configuration.}
We train the model using the Adam optimizer with a learning rate of $10^{-3}$ and weight decay of $10^{-4}$. The models are trained for 50 epochs with a training batch size of 256 and a testing batch size of 64. For the in-domain setting, we split the cached snippet dataset into 80\% for training and 20\% for validation.

\vspace{1mm}
\noindent\textbf{MIL Hyperparameters.}
To aggregate snippet-level logits into a video-level score, we employ Log-Sum-Exp (LSE) pooling. Unless stated otherwise, no attention/entropy regularization is used (disabled). The total loss is the uncertainty-weighted sum of binary cross-entropy on $\hat{z}_{\text{bag}}$ and the MIL-consistent cosine alignment loss (see next subsection).

\vspace{1mm}
\noindent\textbf{Loss Function.}
The model is optimized using a joint objective combining semantic alignment (Cosine Embedding Loss, $\mathcal{L}_{sim}$) and collision classification (Binary Cross Entropy, $\mathcal{L}_{cls}$ with positive weighting). To balance these objectives without manual tuning, we employ homoscedastic uncertainty weighting. The total loss is computed as:

\begin{equation}
    \mathcal{L}_{total} = \frac{1}{2\sigma^2_{sim}}\mathcal{L}_{sim} + \frac{1}{2\sigma^2_{cls}}\mathcal{L}_{cls} + \log(\sigma^2_{sim}) + \log(\sigma^2_{cls})
\end{equation}

\noindent where $\sigma_{sim}$ and $\sigma_{cls}$ are learnable parameters that dynamically weight the task-specific losses during training.

\vspace{1mm}
\noindent\textbf{MIL-consistent Loss Functions.}
For MIL the loss functions remain very similar. The BCE loss $\mathcal{L}_{cls}$ remains the same, however modifications are needed for the similarity loss $\mathcal{L}_{sim}$. The cosine alignment loss is applied per snippet and then aggregated differently for positives vs. negatives:
\begin{itemize}
    \item Positive videos: We weight snippet-text similarities by the LSE-induced attention (i.e., the soft contribution of each $z_t$ to $\hat{z}_{\text{bag}}$), encouraging the few contributive snippets to align more strongly with the caption.
    \item Negative videos: We uniformly average over snippets to discourage uniformly high activations.
\end{itemize}

\vspace{1mm}
\noindent\textbf{Thresholding via Youden's J Statistic for F1/Accuracy.}
We convert predicted probabilities to labels using a single threshold $\tau$ chosen on the validation set by maximizing Youden’s J:
$$J(\tau)=\mathrm{TPR}(\tau)-\mathrm{FPR}(\tau),\qquad 
\tau^\star=\arg\max_{\tau\in[0,1]} J(\tau).
$$
We then report F1 and Accuracy on the evaluation split at threshold $\tau^{\star}$.

\vspace{1mm}
\noindent\textbf{Downstream Integration.}
For the closed-loop evaluation, we fine-tune the TransFuser++ agent for 5 epochs using the pre-trained checkpoints. Fine-tuning uses all the same hyperparameters as the pre-training configuration, but with a smaller learning rate (1e-4). The VLAAD-predicted risk scores are projected and concatenated as an auxiliary token to the driver's input sequence.

\subsection{Additional CARLA-Domain Ablation for VLAAD-MIL}

To further assess the robustness of the CARLA-domain VLAAD-MIL configuration, we vary the amount of normal SimLingo data while keeping the anomaly set fixed to \texttt{carla\_aug} (Carla-Collide, 5k samples). Here, \texttt{simX} denotes an $X$k-sized subset of normal driving clips, and \texttt{carla\_aug} denotes the CARLA-Collide anomaly set used in our in-domain MIL study. All experiments use the same architecture and optimization settings as Section~B.2, including LSE pooling with $\gamma = 10.0$, and are evaluated on the CARLA validation split ($n=425$) over five random seeds.

Table~\ref{tab:carla_scaling} shows a clear data-scaling trend. Increasing the normal set from \texttt{sim5k} to \texttt{sim10k} substantially improves AUC, but gains saturate thereafter: \texttt{sim15k} remains within error bars of \texttt{sim10k}, while larger subsets do not provide consistent improvement. Taken together, these results indicate that in this CARLA-domain MIL setting, approximately 10k normal clips are sufficient to realize most of the attainable AUC improvement.

\begin{table}[t]
\centering
\small
\caption{Effect of scaling the amount of normal SimLingo training data while keeping the anomaly set fixed to \texttt{carla\_aug}. All models use the same VLAAD-MIL architecture and optimization settings as Section~B.2, including LSE pooling with $\gamma=10.0$, and are evaluated on the CARLA validation split ($n=425$). Results are reported as mean $\pm$ std over 5 random seeds. Performance improves substantially from \texttt{sim5k} to \texttt{sim10k}, then saturates; larger normal sets provide no consistent AUC gain.}
\label{tab:carla_scaling}
\begin{tabular}{lcc}
\toprule
Normal subset & AUC $\uparrow$ & Train time (s) \\
\midrule
\texttt{sim5k} & 0.7048 $\pm$ 0.0157 & 173 \\
\textbf{\texttt{sim10k}} & \textbf{0.7373 $\pm$ 0.0138} & \textbf{229} \\
\texttt{sim15k} & 0.7361 $\pm$ 0.0296 & $\sim$270 \\
\texttt{sim20k} & 0.7104 $\pm$ 0.0088 & $\sim$310 \\
\texttt{sim25k} & 0.7347 $\pm$ 0.0178 & $\sim$360 \\
\bottomrule
\end{tabular}
\end{table}

\subsection{LLaVA-Next Fine-Tuning}

For a fair and consistent comparison with large vision–language models, we fine-tune LLaVA-Next using a parameter-efficient LoRA strategy while strictly matching the preprocessing and data construction procedures used in our main experiments. All video data from BDD-X and MMAU are first grouped into fixed 10-second clips containing 40 frames, following the temporal settings used in our main experiments.

Because LLaVA-Next is originally trained on single-image inputs, each video clip is converted into a multi-image sequence by inserting multiple \texttt{<image>} tokens corresponding to the 40 preprocessed frames. These sequences are transformed into LLaVA’s conversational format by pairing the image tokens with the text annotations provided by BDD-X and MMAU. For BDD-X, we use the human-written causal explanations accompanying each clip. For MMAU, we use the infraction descriptions that specify the type of unsafe behavior and its visual cues. No additional text generation, filtering, or synthetic augmentation is performed, ensuring that LLaVA-Next receives the same information content as the backbone model used in our method.

During fine-tuning, the entire vision encoder of LLaVA-Next is frozen. We train only (i) the multimodal projector, which maps CLIP image embeddings to the language model space, and (ii) LoRA adapters inserted into the attention and feed-forward layers of the underlying LLM. This matches the lightweight adaptation philosophy of our own method and avoids overfitting to the limited domain of driving videos. LoRA modules are applied to the query, key, value, output, and MLP layers, allowing efficient low-rank updates while keeping the base model intact.

Fine-tuning is performed on the combined BDD-X and MMAU datasets, using the same batching, sampling, and clip-level formatting used in our model’s comparison experiments. Training objectives follow standard instruction-tuning practice: given a sequence of frames and the corresponding user prompt, the model is optimized with cross-entropy loss on the next-token prediction of the ground-truth response. The following prompts were chosen for different datasets:

\begin{itemize}
    \item For BDDX: \textit{What is happening in this driving scene? Provide a concise description of the scene, including the key agents and their actions.}
    \item For MM-AU: \textit{Is this scene dangerous? If yes, describe the anomaly or unsafe event visible in the clip.}
\end{itemize}

For evaluation, because LLaVA-Next is generative, not a native classifier, we convert its textual outputs into clip-level danger predictions using a structured prompting and scoring procedure. At inference time, each clip is presented using the same prompt used during fine-tuning:
\textit{<image> Is this scene dangerous? Answer yes or no, and briefly explain.}
This prompt exactly matches the natural-language probing protocol used in the paper’s evaluation of scene danger and is consistent with the prompts used during data creation. We extract from the model the token-level probabilities assigned to the first non-image output token being either \textit{yes} or \textit{no}, and use it as the raw score to calculate the entries reported in Table 2 (AUC, F1, Accuracy).

\section{More information on Related Works}
\label{sec:lit_review_app}

\subsection{Collision Data Generation}

Recent efforts in synthetic collision data generation aim to create diverse, safety-critical scenarios in simulation. DeepAccident~\cite{wang2024deepaccident} and  DiffScene~\cite{diffscene} are representative examples that are designed to generate accident scenarios reproduced in CARLA using structured templates derived from NHTSA crash reports. These datasets are exclusively restricted to intersection-based collisions, significantly limiting the coverage of broader vehicle–pedestrian–object interactions and naturalistic driving behaviors. Furthermore, they provide only coarse annotations, making it less suitable for modern vision-language frameworks that demand fine-grained contextual cues in the form of natural language.

Other simulation frameworks, such as 3CSim~\cite{cavojsky20243csim} and Anovox~\cite{bogdoll2024anovox}, focus on generating anomalous and corner cases in driving scenes. However, the temporal driving anomalies they produce are very limited, and lack the diversity required for testing real-world driving conditions.
Despite these advancements, existing simulation datasets frequently fail to span the full spectrum of collision patterns necessary for robust temporal risk modeling. Our CARLA-Collide dataset addresses this critical gap through large-scale, automatic generation of diverse collisions, enriched with multimodal annotations and VLM-ready textual descriptions.

\subsection{VLMs for Road Risk Analysis}

VLMs offer a unified, interpretable framework for understanding complex driving scenes by aligning visual content with rich textual semantics. Models such as CLIP~\cite{CLIP} and XCLIP~\cite{XCLIP} have demonstrated strong semantic reasoning capabilities, enabling robust scene understanding without the need for handcrafted perception pipelines. Several recent works have explored leveraging VLMs for driving tasks. For instance, Think-Driver~\cite{ThinkDriverVLM} uses a VLM to produce high-level driving intentions and risk assessments; however, its interaction with the simulator operates in a benchmark-style loop where commands are issued at slow, non–real-time frequencies and executed by external simulator logic rather than a fully closed-loop controller. Another recent work, SimLingo~\cite{simlingo}, is a vision-language-action (VLA) model that performs closed-loop driving in CARLA but relies on large, slow models and chain-of-thought-style reasoning. Its collision understanding emerges implicitly through its language–action alignment rather than explicit modeling of collision representations.
While these systems confirm the promise of multimodal reasoning for autonomous driving, most operate either in slow command-generation regimes or focus on general scene understanding rather than precise, early-stage collision localization. In contrast, our framework is specifically designed to generate timely, fine-grained collision risk signals over temporally extended clips. By coupling a frozen XCLIP backbone with lightweight adapters and a multiple-instance learning structure, our method explicitly models collision risk in E2E driving stacks which require fast, reliable collision indicators rather than high-latency instruction-level reasoning.
\end{document}